\documentclass{article} 
\usepackage{colm2024_conference}

\usepackage[utf8]{inputenc}
\usepackage[T1]{fontenc}
\usepackage{amsmath}
\usepackage{amssymb}
\usepackage{amsfonts}
\usepackage{graphicx}
\usepackage{booktabs}
\usepackage{multirow}
\usepackage{tabularx}
\usepackage{caption}
\usepackage{subcaption}
\usepackage{enumitem}
\usepackage{xcolor}

\usetikzlibrary{decorations.pathreplacing, calc, positioning, arrows.meta, fit}

\newcommand{\rewrite}[1]{\textcolor{red}
{\textbf{[REWRITE PENDING]}}}

\title{\textbf{Ovis-Embedding: Pushing the Frontiers of Universal Omni-Modal Embeddings}}

\author{
Ovis-Embedding Team \\[2mm]
\textbf{Alibaba Token Hub, Alibaba Group}
}

\begin{document}

\maketitle

\begin{abstract}

In this report, we introduce \textbf{Ovis-Embedding}, a state-of-the-art omni-modal embedding family built on native integration of text, image, video, and audio. Instead of assembling separate modality towers, Ovis-Embedding uses a shared multimodal backbone to encode different modalities in a common representation space. Specifically, we make \textbf{three key advances}: (1) \textbf{native omni-modal initialization}: we adopt a pretrained Qwen-omni model as the embedding backbone and adapt it through contrastive training with low-rank initialization; (2) \textbf{data-centric omni-modal training}: we construct a broad, high-quality corpus spanning text, images, video, audio, and interleaved multimodal data. To improve data efficiency, we introduce homogeneous-source sampling to form task-consistent batches with informative in-batch negatives; and (3) \textbf{embedding-specific training and inference optimization}: we use focal loss to emphasize hard examples and similarity-based Embedding Distillation to transfer fine-grained similarity structure from complementary experts. At inference time, low-rank feature decomposition enables compact embeddings with flexible dimensionality and minimal performance loss. Empirical evaluations show that the \textbf{Ovis-Embedding} family achieves state-of-the-art performance on \textbf{MMEB-v3}, \textbf{MMEB-v2}, \textbf{MVEB}, \textbf{MAEB}, and \textbf{RTEB}, demonstrating its effectiveness across text, image, video, and audio modalities. These results highlight the potential of unified omni-modal training to overcome modality fragmentation and advance universal embedding models for any-to-any retrieval.

\end{abstract}

%

\definecolor{Cone}{HTML}{4E79A7}    
\definecolor{Ctwo}{HTML}{F28E2B}    
\definecolor{Cthr}{HTML}{59A14F}    
\definecolor{Cfour}{HTML}{B07AA1}   
\definecolor{Cfive}{HTML}{E15759}   

\providecommand{\teaserbar}[4]{%
  \pgfmathsetmacro{\xL}{#1 + #2*\barW}%
  \pgfmathsetmacro{\xR}{\xL + \barW}%
  \pgfmathsetmacro{\hY}{#3*\unitY}%
  \fill[#4] (\xL,0) rectangle (\xR,\hY);%
  \draw[#4!45!black, line width=0.3pt] (\xL,0) rectangle (\xR,\hY);%
  \draw[white, opacity=0.55, line width=0.5pt]
        (\xL+0.05,\hY-0.06) -- (\xR-0.05,\hY-0.06);%
  \node[font=\tiny\bfseries\sffamily, text=#4!30!black, anchor=west,
        rotate=90, inner sep=1pt]
        at (\xL+\barW/2, \hY+0.04) {#3};%
}

\providecommand{\teaserlegend}[3]{%
  \fill[#1] (#3,\legY) rectangle ++(0.32,0.20);%
  \draw[#1!50!black, line width=0.3pt] (#3,\legY) rectangle ++(0.32,0.20);%
  \pgfmathsetmacro{\labx}{#3+0.40}%
  \node[font=\scriptsize\sffamily, anchor=west]
        at (\labx, \legY+0.10) {#2};%
}

\begin{figure*}[h]
\centering
\includegraphics[width=0.9\textwidth]{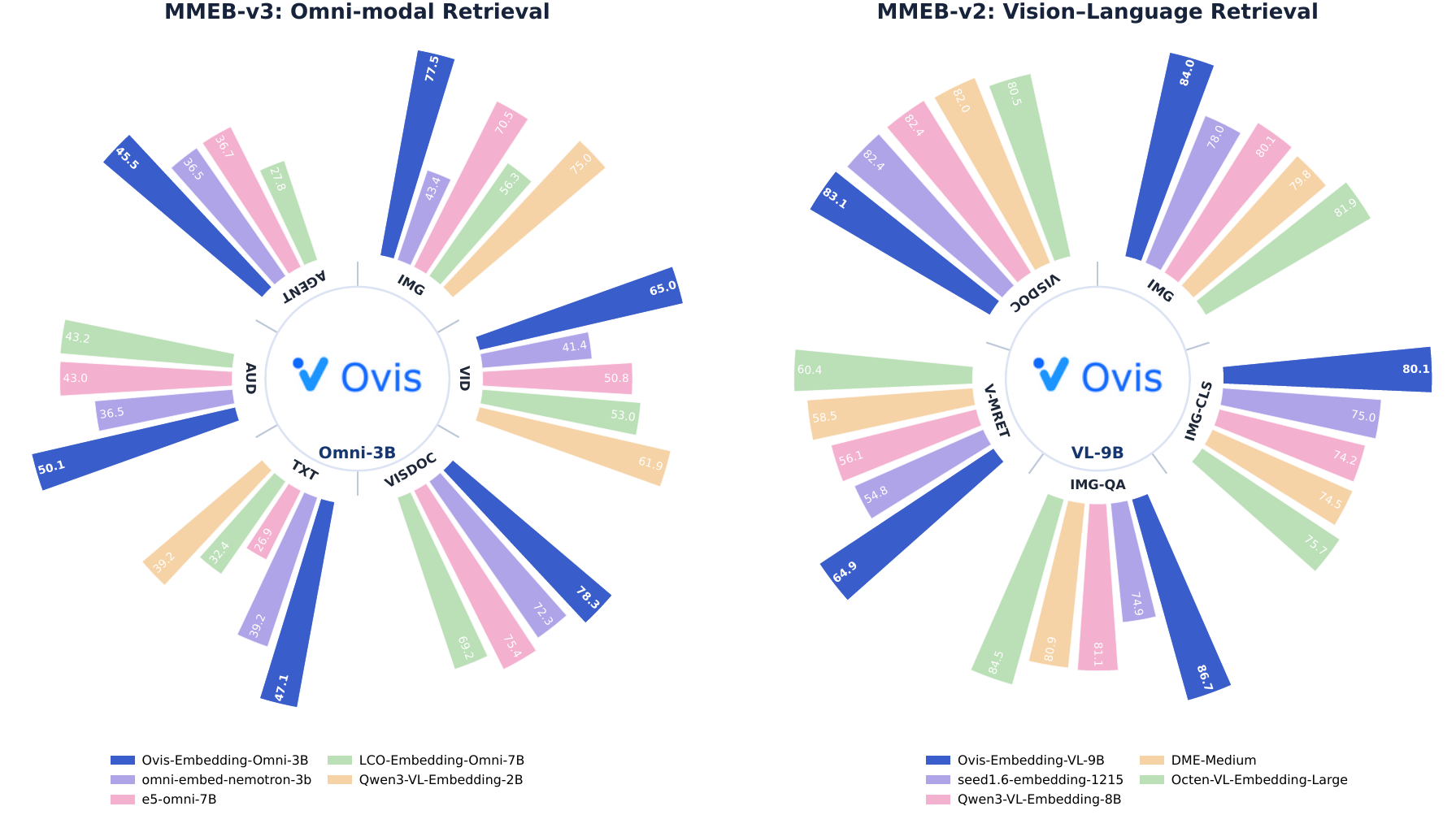}
\caption{\textbf{Ovis-Embedding performance across modalities.}
Ovis-Embedding-Omni-3B leads all six MMEB-v3 modality groups (left), while
Ovis-Embedding-VL-9B leads four of five MMEB-v2 task groups (right). Bar lengths
are normalized within each benchmark; labels show the original percentage
scores.}
\label{fig:architecture}
\end{figure*}
\newpage
\section{Introduction}
\label{sec:intro}

Dense embeddings serve as a core retrieval layer for search,
retrieval-augmented generation, recommendation, and agentic systems~\citep{qwen3embedding,rag,youtube_recs,generative_agents}. As modern information repositories increasingly combine text,
images, videos, audio, visual documents, and interface states, retrieval can no
longer be reduced to a small number of predefined modality pairs. Consider a
maintenance agent investigating an abnormal machine sound: it may use an audio
clip together with a short textual description to retrieve a relevant video
tutorial, a diagram in a PDF manual, or a previous service record. Solving such
a task requires a query---potentially composed of multiple modalities---to be
matched directly against candidates in different modalities within the same index~\citep{mm_embed,omni_embed_nemotron}. Merely representing heterogeneous inputs as vectors of
equal size is insufficient. Their representations must instead form a coherent,
calibrated semantic space in which relevance scores are comparable across
modalities while the fine-grained distinctions needed within each modality are
retained~\citep{mmebv3}. We refer to this general retrieval setting as
\emph{any-to-any retrieval}.

Existing multimodal embedders provide an incomplete foundation for universal retrieval. Vision--language specialists~\citep{qwen3_vl_embedding,vlm2vec_v2} do not support audio, while existing omni-modal systems~\citep{e5_omni,lco_embedding,conan_embedding,jina_v5_omni} typically add an audio pathway to a pretrained text--vision embedder or align a separate audio encoder with an established embedding space. In both cases, acoustic inputs are adapted to a geometry learned without them, which may limit fine-grained alignment across modalities. We introduce \textbf{Ovis-Embedding}, built directly on the pretrained Qwen-Omni understanding model~\citep{qwen25omni}, whose text, vision, audio, and video inputs are processed by a shared model. We remove the speech-generation pathway and use the final-layer state at the last non-padding token as the embedding, without adding modality-specific projection heads. This converts the native understanding backbone into a unified encoder for any-to-any retrieval with either unimodal or interleaved inputs.

To train this representation, we build a large-scale corpus spanning text, images, video, audio, and interleaved inputs. Text data cover retrieval and semantic matching. Image data support classification, question answering, retrieval, and grounding, while video data consist of query--video pairs verified by a VLM. Audio data cover recognition and bidirectional audio--text retrieval. Interleaved samples combine multiple modalities, and agent data target tool, GUI, and knowledge retrieval. We cast samples as query--positive--negative tuples, discard invalid or duplicated items, filter false negatives, and balance query and candidate modalities. All multimodal data are then fed into our omni-embedding model.


We develop a coordinated optimization-and-deployment recipe that progressively establishes broad omni-modal alignment. We begin with low-rank contrastive pretraining on the large-scale omni-modal corpus, gathering candidates across data-parallel workers to form a broad cross-modal negative pool. A difficulty-aware focal objective concentrates learning on unresolved queries, while precomputed similarity distributions from modality experts provide graded ranking supervision over both positive and negative candidates. We then unfreeze the full model and refine it on higher-quality data using the homogeneous-source batches described above, sharpening discrimination among task-consistent candidates. As optimization approaches saturation, \textbf{Embedding Distillation} retains teacher-correct examples, upsamples cases still missed by the student, and assigns stronger ranking supervision to less confident queries, transferring expert capabilities without introducing modality-specific components at inference. Finally, a low-rank feature transformation with lightweight residual adaptation produces multiple compact embedding dimensions from the same encoder, reducing index storage and similarity-computation costs with minimal loss in retrieval quality.

Together, the native initialization, comprehensive data pipeline, and embedding-specific training and inference optimizations produce state-of-the-art results across both general and modality-intensive evaluation: \textbf{Ovis-Embedding-Omni-3B} establishes a new state-of-the-art on \textbf{MMEB-v3},
\textbf{Ovis-Embedding-VL-9B} ranks top on \textbf{MMEB-v2},
and the Ovis-Embedding family further advances \textbf{MVEB}, \textbf{MAEB}, and \textbf{RTEB}. To make these advances broadly accessible, we will open-source the model checkpoints, training and data-construction recipes, inference code, and unified evaluation toolkit. We hope this will provide the community with a foundation for developing and evaluating universal embedding models, particularly for the audio, audio--video, and any-to-any retrieval settings that remain under-served by existing open resources.

\paragraph{Contributions.}
Our work makes the following contributions.
\begin{itemize}
    \item \textbf{Native omni-modal initialization for universal embeddings.}
        Unlike existing approaches that retrofit an audio branch onto a
        vision--language embedder or align a separately trained audio embedding
        model, we start directly from a pretrained omni-modal understanding model.
        By retaining Qwen-Omni's natively aligned text, vision, audio, and video
        front-ends and their shared encoder, Ovis-Embedding learns any-to-any
        retrieval in a coherent representation space without modality-specific
        embedding heads.

  \item \textbf{Omni-modal homogeneous-source sampling.}
      We introduce a unified sampling strategy across text, image, video, audio,
      visual-document, agent, and interleaved multimodal tasks. Drawing each
      micro-batch from one source and deduplicating pooled candidates produces
      task-consistent hard negatives, reducing modality shortcuts and improving
      fine-grained discrimination in the shared embedding space.

    \item \textbf{Embedding-specific training and inference optimization.}
        During training, difficulty-aware focal loss bootstraps a robust universal
        embedding space by emphasizing unresolved hard examples, while
        similarity-based Embedding Distillation transfers the fine-grained
        similarity geometry of complementary experts into a single encoder. At
        inference time, low-rank feature decomposition supports compact embeddings
        with flexible dimensionality, reducing retrieval storage and computation
        with minimal performance loss.


    \item \textbf{Open state-of-the-art models for the community.}
        We will release a family of state-of-the-art checkpoints covering both
        vision--language and native omni-modal retrieval, together with training
        and inference code and a unified evaluation toolkit. Ovis-Embedding-Omni-3B
        establishes a new state of the art on MMEB-v3, Ovis-Embedding-VL-9B leads
        MMEB-v2, and the family further advances the state of the art on MVEB,
        MAEB, and RTEB, providing an accessible foundation for universal embedding
        research.
\end{itemize}

\begin{figure*}[t]
\centering
\includegraphics[width=1.0\textwidth]{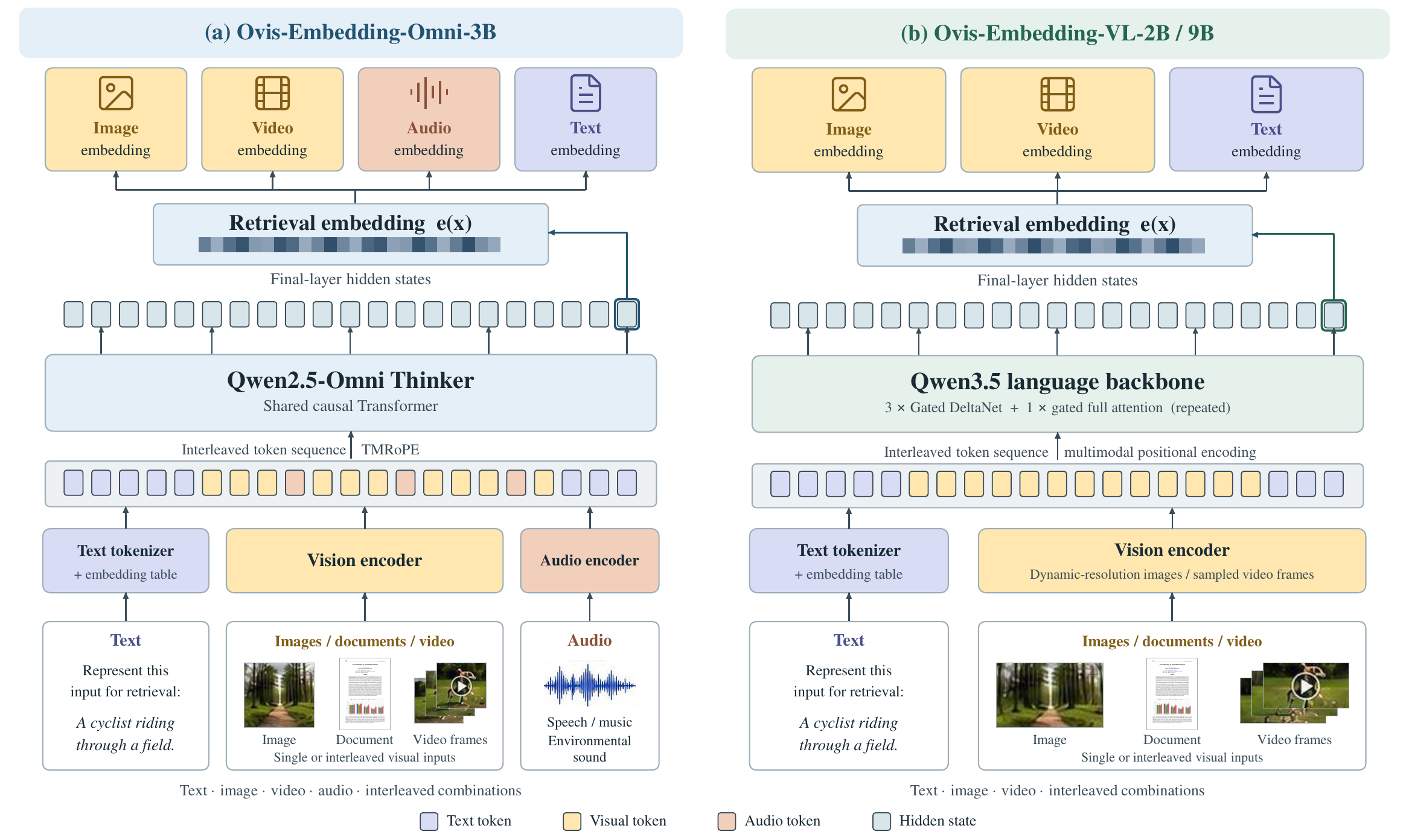}
\caption{\textbf{Architecture of Ovis-Embedding.} (a) Ovis-Embedding-Omni-3B
encodes text, visual, and audio inputs as an interleaved token sequence processed
by the Qwen2.5-Omni Thinker with TMRoPE. (b) Ovis-Embedding-VL-2B/9B processes
text and visual inputs with the Qwen3.5 language backbone. Both variants use the
final-layer hidden state at the last non-padding token as the retrieval
embedding, supporting both unimodal and interleaved inputs without
modality-specific projection heads.}
\label{fig:architecture}
\end{figure*}

\section{Model Architecture}
\label{sec:architecture}

\paragraph{Backbone family.}
Ovis-Embedding is instantiated from two complementary native multimodal
backbones. \textbf{Ovis-Embedding-Omni-3B} is initialized from
Qwen2.5-Omni-3B~\citep{qwen25omni} and supports text, images, video, and audio.
\textbf{Ovis-Embedding-VL-2B} and \textbf{Ovis-Embedding-VL-9B} are initialized
from Qwen3.5-2B and Qwen3.5-9B, respectively, and support text,
images, and video. Thus, all three models start from backbones that were trained
to fuse multiple modalities natively; the distinction is that the Omni variant
additionally provides a native audio pathway, whereas the VL variants devote
their capacity to vision--language representation learning. This design gives
the model family a common retrieval interface while covering different
modality, accuracy, and efficiency requirements. The overall architecture is shown in Fig.~\ref{fig:architecture}.

\paragraph{Embedding adaptation.}
We convert each generative backbone into a bi-encoder using the same minimal
adaptation. An input $x$, which may contain one modality or an interleaved
combination of supported modalities, is formatted with a task instruction by
the backbone's native processor and chat template. The resulting text and
modality tokens are processed jointly by the pretrained backbone. We remove the
language-modeling output head and use the final-layer hidden state at the last
non-padding token as the sequence representation,
\begin{equation}
    \mathbf{e}(x) = \mathbf{h}^{(L)}_{\ell(x)} \in \mathbb{R}^{d},
\end{equation}
where $L$ is the number of backbone layers, $\ell(x)$ denotes the last valid
token position, and $d$ is the native hidden size of the selected backbone. We
do not introduce an additional embedding projection or modality-specific output
head. Consequently, the embedding dimensionality is inherited directly from
the backbone, and all supported input types are mapped through the same output
interface. Training and retrieval use cosine similarity (equivalently, a dot
product between $\ell_2$-normalized embeddings), as defined in
Section~\ref{sec:objective}. This parameter-free conversion preserves the
cross-modal alignment learned during multimodal pretraining while specializing
the representation space for retrieval.

\subsection{Omni Architecture}
\label{sec:omni-architecture}

The Omni model uses the \emph{Thinker--Talker} architecture of
Qwen2.5-Omni-3B~\citep{qwen25omni}. In the original model, a vision encoder maps
images and video frames to visual tokens, an audio encoder maps speech, music,
and environmental sounds to acoustic tokens, and a tokenizer provides text
tokens. These streams are interleaved and consumed by the shared causal
Transformer, termed the \emph{Thinker}. Time-aligned Multimodal Rotary Position
Embedding (TMRoPE) aligns the temporal positions of audio and video, allowing
the Thinker to model synchronized audio--visual content in addition to
single-modality inputs. The separate \emph{Talker} predicts streaming speech
units from Thinker representations in the original generative system.


For Ovis-Embedding-Omni-3B, speech generation is unnecessary. We therefore
discard the Talker and retain the text tokenizer, vision encoder, audio encoder,
and Thinker. The pooled Thinker state is used directly as the embedding. Unlike
approaches that attach independently trained modality towers to a text model,
this construction reuses a backbone in which text, image, video, and audio were
already aligned through a shared Transformer. Consequently, each side of a
retrieval pair---both the query and the candidate---may contain any combination
of text, images, video, and audio, including interleaved and synchronized inputs.
Conventional unimodal or pairwise retrieval, such as text--text, image--text,
video--text, audio--text, and audio--video retrieval, therefore becomes a special
case of general any-to-any retrieval within one representation space.

\subsection{Vision--Language Architecture}
\label{sec:vl-architecture}

The VL models use the dense Qwen3.5-2B and Qwen3.5-9B backbones~\citep{qwen3.5}.
Qwen3.5 is natively trained on interleaved text, image, and video tokens rather
than extending a text-only model with a post-hoc retrieval tower. Its vision
encoder converts dynamic-resolution images and temporally sampled video frames
into compact visual-token sequences, which are inserted into the language-token
stream and processed jointly by the causal backbone. Multimodal rotary position
encoding preserves temporal and two-dimensional spatial coordinates for these
visual tokens.

The Qwen3.5 language backbone uses a hybrid stack with three Gated DeltaNet
linear-attention layers for every full gated-attention layer. This design uses
linear attention for efficient long-context processing while periodically
applying full attention for precise global token interaction. Qwen3.5-2B uses
24 backbone layers with hidden size 2,048, whereas Qwen3.5-9B uses 32 layers with
hidden size 4,096. We remove the language-modeling output head from both models
and apply the shared last-token pooling rule described above, without adding a
projection head. The 2B variant provides a compact vision--language embedder,
while the 9B variant provides higher capacity; both retain a common training
objective and inference interface for text, image, video, and their interleaved
combinations.

\section{Training Data}

\begin{figure*}[t]
\centering
\includegraphics[width=1.0\textwidth]{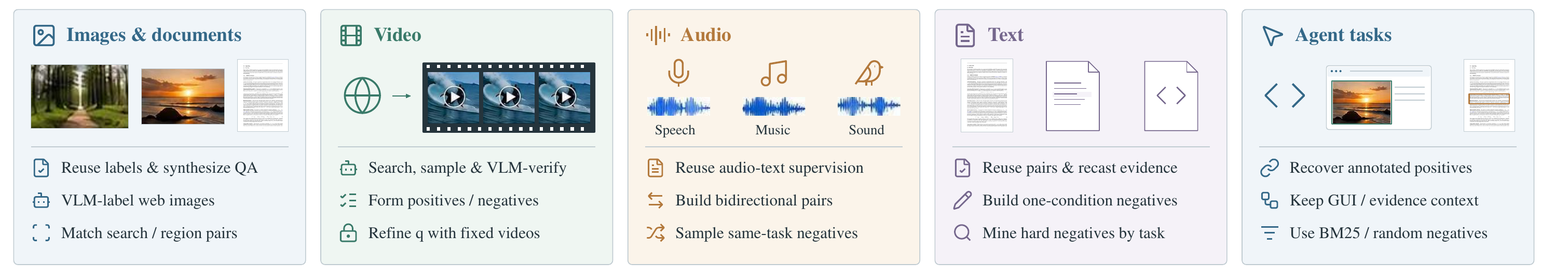}
\caption{\textbf{Modality-specific data construction.} Training pairs for
images and documents, video, audio, text, and agent tasks are constructed with
task-specific supervision, semantic verification, pair formation, and negative
mining. These pipelines produce quality-controlled examples for unified
omni-modal embedding training.}
\label{fig:data}
\end{figure*}

We construct a large-scale training corpus spanning text, images and visual
documents, video, audio, and interleaved multimodal inputs from both public and
proprietary sources. Figure~\ref{fig:data} summarizes the
modality-specific construction pipelines. For images and documents, we reuse
annotated datasets, synthesize question--answer pairs, label web images with a
VLM, and form search and region-level pairs. For video, we retrieve web clips,
verify their sampled frames with a VLM, construct positive and negative pairs,
and refine queries while keeping the selected videos fixed. For audio, we reuse
audio--text supervision to build bidirectional pairs with same-task negatives.
For text, we recast existing pairs and evidence as retrieval examples and
construct task-specific hard negatives, including negatives that violate a
single query condition. For agent tasks, we recover annotated positives,
preserve GUI and evidence context, and use BM25-based or random negatives.

All examples are standardized as query--positive--negative tuples and undergo
deduplication and multi-stage quality filtering. All data associated with
evaluation test sets are strictly deduplicated against the training corpus. For
this audit, we use normalized text matching for textual data and perceptual-hash
matching for visual data, removing all detected overlaps. The resulting corpus
contains approximately \textbf{50M}
(query, target) training pairs.\footnote{All numbers reported in this subsection
are placeholder estimates based on the current data freeze; the final figures
will be updated for the camera-ready version.}

\subsection{Image Data}

We organize the image training corpus around four complementary task families:
image classification, image question answering, image retrieval, and image
grounding (GD). The data format can be found in Fig.~\ref{fig:image-data-1} and Fig.~\ref{fig:image-data-2}

\textbf{Image classification.} We draw the core supervision from classical
computer-vision datasets, including ImageNet~\citep{imagenet}, CUB-200~\citep{cub_200}, and SUN397~\citep{sun397}, which
collectively cover generic objects, fine-grained bird species, and diverse scene
categories. To extend this supervision beyond the closed taxonomies and visual
distributions of established benchmarks, we further collect candidate images
returned by web search and use Qwen3.5-Plus to inspect their visual content and
assign category labels. The resulting web-augmented classification data
substantially broaden both category coverage and intra-class appearance
variation.

\textbf{Image question answering.} We construct and synthesize training
examples from established sources such as TextVQA~\citep{textvqa}, DocVQA~\citep{docvqa}, and InfoVQA~\citep{infographicvqa}, covering scene text, document understanding, and information-rich visual content. We complement these conventional tasks with synthetically generated
QA data for long-tail scenarios that are sparsely represented in public
datasets, thereby improving coverage of uncommon entities, specialized
contexts, and less frequent visual reasoning patterns.

\textbf{Image retrieval.} The data are collected from three principal sources:
news image--text pairs, text-to-image search data, and image-to-image search
data. News data provide semantically rich correspondence between visual events
and their textual context, text-to-image search captures open-domain user intent
expressed in natural language, and image-to-image search supplies direct
supervision for instance-level and semantic visual matching.

\textbf{Image grounding.} We construct region-aware supervision from MS-COCO~\citep{coco}
and its subsequent derivative datasets, converting their object, region, and
language annotations into fine-grained grounding examples.

\textbf{Open-world image retrieval.} Beyond these four standard task families,
we also collect concrete queries issued by online users and use Quark web search
to retrieve relevant content and synthesize a broader product-oriented dataset.
This additional data introduces realistic, colloquial, and highly diverse
shopping intents, extending the image corpus from benchmark-defined tasks to
practical open-world retrieval. Together, these sources provide supervision at
category, question-answering, global retrieval, and local grounding levels,
enabling the model to learn both broad visual semantics and fine-grained
cross-modal relevance.

\subsection{Video Data}

We adopt a general collect--filter--organize pipeline
to construct video data for a broad range of multimodal embedding tasks.
Task-relevant textual descriptions are used to retrieve an initial pool of web
videos with broad visual coverage. The collected candidates are subsequently
filtered and organized through content-based validation tailored to each
downstream task. The downloader paginates through search results, applies basic
duration and format constraints, prefers suitable video encodings, retries
failed downloads, and records provenance and technical metadata. The resulting
candidates are treated as noisy, unlabeled videos until their visual content has
been independently validated.

\textbf{Semantic validation.} The validation procedure is adapted to the target
task. Uniformly sampled frames from each candidate are provided to a
vision--language model together with a task-specific instruction. The model
determines whether the observable content satisfies the intended semantic
condition and returns a structured judgment. Only candidates passing this
content-based verification are converted into training examples. This shared
framework can support classification, retrieval, and other video understanding
objectives by changing the discovery prompts, validation criteria, and final
positive--negative organization.

\textbf{Video classification.} Candidate discovery is organized around a broad
vocabulary of actions and events relevant to the desired semantic coverage.
Importantly, these concepts act only as retrieval cues during acquisition. For
every candidate clip, a vision--language model independently verifies whether
the sampled visual content actually depicts the corresponding action or event.
The discovery concept is converted into a positive training label only after
this verification succeeds; unrelated or ambiguous search results are
discarded. Each accepted video is then paired with its verified category text,
while category descriptions sampled from the remaining task vocabulary serve as
negative texts. Thus, the final classification supervision is determined by
video-content validation rather than by the ranking or metadata returned by the
search engine.

\textbf{Video retrieval.} Natural-language descriptions are used to retrieve an
initial pool of candidate videos. The subsequent processing follows a
filter-then-refine procedure. First, a vision--language model evaluates each
downloaded video against its original query and assigns an ordinal relevance
score based on uniformly sampled frames. Only query--video pairs receiving the
highest relevance level are retained. These filtered pairs are then organized
into contrastive samples: the retained video serves as the positive candidate,
while negative videos are randomly sampled from other retained pairs associated
with different normalized query strings.

After the positive and negative candidate sets have been fixed, a second
vision--language model pass performs conservative, video-grounded query
refinement. The model receives the original query together with frames sampled
from the verified positive video and determines whether the text accurately
describes the visible content. If the query is already accurate, it is preserved
unchanged. Otherwise, the model corrects only observable factual mismatches,
such as the number of people, the depicted action, or the involved objects. The
revised query is required to remain concise, preserve the correct portions and
expression style of the original text, and avoid introducing details that
cannot be clearly observed. The refinement result is matched back to the sample
through the positive-video identity, and only the query text is updated; the
previously selected positive and negative videos remain unchanged. This ensures
that query rewriting is applied only to video--text pairs that have already
passed the relevance filter, rather than directly modifying the raw search
queries or unverified candidates.

Following the filter-then-refine procedure, the finalized query is matched back
to its corresponding sample through the positive-video identity. Only the
textual supervision is updated, while the positive and negative candidate sets
established during the filtering stage remain unchanged. The referenced videos
are subsequently converted into a consistent input representation, and
incomplete samples are removed through basic integrity checks. Overall, this
pipeline transforms broadly retrieved video candidates into content-grounded
multimodal examples through task-specific validation and conservative semantic
correction. The classification and retrieval cases illustrate how the same
general construction framework can be adapted to different learning
objectives: classification data emphasize verified video--category alignment,
whereas retrieval data preserve fine-grained video--text correspondence through
relevance filtering and query refinement. The framework can be extended to
additional video tasks by modifying the semantic validation criteria and the
organization of positive and negative supervision.

\subsection{Audio Data}
\label{sec:audio-data}

We construct the audio training corpus to cover three complementary forms of
acoustic understanding. First, recognition-oriented data associate audio clips
with semantic categories spanning environmental events, urban sounds, musical
instruments, and spoken commands. Second, audio--language matching data pair
non-speech sounds with natural-language descriptions and speech segments with
their transcripts. Third, bidirectional retrieval data train both audio-to-text
and text-to-audio matching, allowing either modality to serve as the query.
Together, these sources expose the model to speech, music, and diverse everyday
sound events rather than allowing the audio representation to be dominated by a
single acoustic domain.

\textbf{Audio recognition.} The recognition data provide supervision at several
levels of semantic granularity. Broad sound-event and acoustic-scene labels
teach the encoder to identify the dominant source and context of a recording,
while instrument and spoken-word labels emphasize timbre, phonetic content, and
short-duration cues.

\textbf{Audio--language matching.} The descriptive data complement these
relatively compact label spaces with free-form captions that may refer to
multiple co-occurring events, their temporal evolution, and the surrounding
acoustic environment. Speech segments paired with transcripts further connect
linguistic content expressed in audio to its textual realization. Combining
these forms of supervision encourages the model to preserve both non-linguistic
acoustic semantics and spoken-language information in the same representation.

\textbf{Unified contrastive format.} We convert every source into a shared
contrastive-retrieval format consisting of a query, a matched target, and
task-consistent negative candidates. For recognition tasks, an audio query
retrieves its corresponding textual category. For caption and transcription
tasks, the target is a natural-language description or transcript with
substantially richer semantics.

\textbf{Bidirectional retrieval.} Where paired audio--text data are available,
we construct both retrieval directions: audio queries retrieve text, and
textual queries retrieve audio. This symmetric construction reduces directional
bias and makes the learned space directly usable for cross-modal search
regardless of which modality initiates retrieval.

\textbf{Negative construction and filtering.} Negative candidates are kept
within the same task family whenever possible, so that the model must
distinguish semantically related sounds or descriptions rather than relying on
coarse differences in input format. Before training, we apply basic format and
content validation, remove invalid or duplicate entries, and balance
heterogeneous sources to limit domination by the largest speech or caption
collections. The resulting corpus can be mixed with text, image, and video
supervision under the same embedding objective, requiring neither an
audio-specific loss nor a separate audio output space.

\subsection{Text Data}

Following the fine-grained text retrieval categories proposed by~\citet{vlm2vec}, we construct text datasets across
five primary task paradigms. What separates these paradigms is not their subject matter but the extra relevance criterion they impose and
consequently the locus at which a training signal has to be manufactured.

\paragraph{General text retrieval.}
The query $q$ expresses an information need over a passage collection $\mathcal{C}$, and $d^{+}\in\mathcal{C}$
is a passage judged to address that need; no additional criterion is imposed. Relevance is therefore a single
unstructured judgment that does not decompose into separately checkable terms. The scarce resource in this
regime is the positive, since relevance judgments are human-annotated and cannot be manufactured. We
therefore build on the established supervised collections for dense passage retrieval and treat this paradigm
as the semantic substrate for text embedding.

\paragraph{Instruction-following retrieval.}
The query is a composition $q=x\oplus c$ of an underlying request $x$ and a natural-language constraint $c$,
typically bearing on audience, clarity, format, language, length, or source. Relevance is conjunctive: writing
$d\models\varphi$ for the judgment that $d$ satisfies a stated condition $\varphi$,
\begin{equation}
\mathrm{rel}(q,d)\;\Longleftrightarrow\;(d\models x)\wedge(d\models c),
\end{equation}
and since the request term is satisfied by many candidates on the same topic, the discriminative signal resides
entirely in the second term. An
ideal hard negative therefore satisfies $d^{-}\models x$ but $d^{-}\not\models c$. With the request held fixed, topical similarity no longer provides a usable cue, compelling the model to represent the constrained attribute 
$c$ in itself, rather than folding it into the query as additional topical keywords. We collect released supervised train
splits on instruction-following tasks, keeping the constraint inline with the request which matches the concatenated form presented at evaluation.

\paragraph{Reasoning retrieval.}
Relevance here demands logical inference and multi-hop reasoning beyond keyword matching: the query $q$ poses a
problem or presents a case, and $d^{+}$ supplies the principle, evidence, or precedent under which it is
resolved, so the pair is linked by a latent inference $r$ with $q\xrightarrow{\;r\;}d^{+}$. The contrast with
general retrieval is that the proposition connecting them---the diagnosis, the applicable theorem---is stated in
neither, so no degree of surface or semantic proximity between $q$ and $d^{+}$ recovers the judgment. We obtain
such pairs by recasting generative reasoning corpora into retrieval form, promoting whichever field realizes
this relation to $d^{+}$: a derivation, a supporting passage, or a case drawn from the same source.

\paragraph{Multi-condition retrieval.}
The query is an ordered conjunction of atomic conditions, $q_k=\langle f_1,\dots,f_k\rangle$,
all of which $d^{+}$ must satisfy jointly. We construct training data by
extending the MultiConIR pipeline~\citep{multiconir}. The conditions are extracted
from a source document $d_{\mathrm{src}}$, so $d_{\mathrm{src}}$ satisfies them by
construction and serves as $d^{+}$ at no annotation cost, while the query verbalizes the
first $k$ of them. For each
$f_j$ we generate a replacement $h_j$ that voids the condition while retaining its salient
terms, redeployed so that they no longer license $q_k$, and substitute it into
$d_{\mathrm{src}}$ in isolation, one negative per condition:
\begin{equation}
d^{+}=d_{\mathrm{src}},\qquad
\mathcal{N}(q_k)=\bigl\{\mathrm{HN}_j\bigr\}_{j=1}^{k},\qquad
\mathrm{HN}_j\mathrel{:=} d_{\mathrm{src}}[f_j\!\to\!h_j].
\end{equation}
Each negative thus departs from the positive in exactly one condition and in nothing else.
Hardness is certified by construction rather than estimated by a scorer: any representation
that compresses a document into a single vector must place $\mathrm{HN}_j$ within a
vanishing margin of $d_{\mathrm{src}}$, and closing that margin is precisely what the
paradigm asks of the model.

\paragraph{Long-context retrieval.}
The task form is an extreme asymmetry, $|d^{+}|\gg|q|$, with context length promoted to an explicit variable:
what is retrieved is an entire document, but what makes it relevant is a short span inside it. Relevance is thus
local while the representation is global, and the capability under test is tolerance of that mismatch---the query--document score must not decay as irrelevant material accumulates
around a fixed piece of evidence. The
demand sharpens wherever one document answers many distinct queries, each keyed to a different span: a
fixed-dimensional embedding then cannot stand for the document by its dominant topic, but must retain numerous
local facts at once and keep them separately addressable. What the task probes is therefore long-context
understanding rather than semantic matching, and the binding constraint is
representational capacity, not discrimination among candidates. Most sub-tasks inherit collections whose documents are natively long; the
remainder are synthetic, with length and evidence position set directly.

\paragraph{Text-retrieval data.}

To strengthen the model's retrieval capability, we incorporate RTEB-related training data. In addition to improving text retrieval, this supervision helps the shared embedding space generalize retrieval capability to images, audio, and video. Following the major specialized domains covered by RTEB~\citep{rteb}, our data span law, finance, programming, and healthcare, with retrieval relations including legal scenario--statute matching, financial question--report retrieval, natural-language problem--code matching, and medical question--answer retrieval. We construct query--document pairs from naturally occurring relations in domain corpora, such as questions paired with expert answers, programming problems paired with solutions, and legal scenarios linked to relevant statutes or cases. Structured records, financial tables, source code, and long documents are converted into self-contained candidate texts while preserving information important for retrieval. Existing queries are retained when suitable; otherwise, they are derived from annotations or conservatively rewritten from the associated content. Candidate pairs are then checked through rule-based filtering and, where necessary, model-assisted semantic validation. After deduplication, we retrieve lexically similar candidates from the same domain to form hard negatives and supplement them with broader samples when needed. All known positives and near-duplicate candidates are excluded from the negative set to reduce false negatives. Finally, task-specific query instructions and a unified candidate format are applied before mixing the data into training.

\paragraph{Text data in the wild.}
To broaden the competence of the shared embedding space beyond retrieval, we further incorporate related training data from~\citet{mteb}. This supervision asks a single representation to support judgments of several kinds at once:
category membership, topical grouping, pairwise equivalence, and graded semantic similarity. We therefore
organize the data by the task types of MTEB rather than by domain, building direct supervision for
classification, clustering, retrieval, and pair classification. Since the training objective is uniformly contrastive, the central problem is
not assembling these tasks but casting each of them into a common query--candidate form without distorting what
it measures.

Classification is recast as retrieval over verbalized labels: each label is rendered as a natural-language
candidate, so the decision becomes a choice within a small closed candidate set rather than the output of a
task head. Clustering data is drawn from a labeled topic hierarchy and stratified over its leaf categories, so
that the sampled subset preserves the shape of the label tree, while the duplicate-question and paraphrase
collections supply positives directly from their annotations. The decisive quality issue in this group is the negative set: a negative that
can be rejected by lexical overlap alone leaves a competent model with almost no loss and therefore no gradient,
however well the positive is annotated. We accordingly mine in-domain hard negatives with BGE-M3~\citep{bge_m3} to
replace the weakest random ones. Known positives and near-duplicates are excluded from the negative set, and
candidate text is screened at the document level against the evaluation collections before mixing.

\subsection{Agent Data}

We construct agent data for tool retrieval, GUI interaction retrieval, and
knowledge retrieval, covering both text and multimodal inputs.

\textbf{Tool retrieval.} Examples pair user requests with tool or API
documentation. Positive tools are identified from relevance annotations,
recorded calls, and tools referenced in annotated plans. Candidate documents
preserve tool names and functional descriptions, together with parameter
specifications and implementation details when available. Both queries and tool
candidates are represented as text, with retrieval instructions attached to
queries.

\textbf{GUI retrieval.} GUI data connect user goals, interface states, and
interaction trajectories through four relations: goal to trajectory, goal to
state, state to state, and trajectory to state. We construct positive pairs from
annotated correspondences in interaction records, preserving the associated
text, screenshots, and their ordering. This produces retrieval examples in
which either side may contain text together with one or more screenshots.

\textbf{Knowledge retrieval.} The data pair questions about scientific
documents with supporting paragraphs. We recover positive passages from
evidence annotations and represent each candidate using its section title
followed by the paragraph text. Distinct annotated evidence passages associated
with the same question are collected together, retaining the document context
needed to distinguish supporting evidence from other passages.

\textbf{Negative construction.} We combine BM25~\citep{bm25} hard-negative mining with
provided negatives and sampling. For tool data, provided negatives are
retained, and additional hard negatives are selected by descending BM25 scores
between queries and candidate documents. Mining prioritizes the same domain or
tool category where applicable. Other functions within the associated toolkit
also serve as distractors when they are not annotated as positives. For evidence
retrieval, we prioritize paragraphs from the same document that rank highly
under BM25 but are not annotated as evidence. Insufficient negative sets are
supplemented from broader candidate pools using deterministic selection or
seeded sampling. GUI data use random negatives, sampled uniformly without
replacement from distinct candidates of the same relation type.

\textbf{Preprocessing.} Duplicate queries are consolidated and candidate pools
are deduplicated while preserving internal whitespace, code formatting, and
media order. Negative construction excludes all known positives for each query,
including alternatives that will not be used as positive supervision. For
entries containing multiple positive samples, only the first positive is used
during training, and the remaining positives are discarded.

\section{Omni-Modal Training}
\label{sec:training}
Our framework comprises four stages. Stage-1 establishes a unified omni-modal embedding space through low-rank pretraining (initialization); Stage~2 refines it through full-parameter finetuning with homogeneous sampling; Stage-3 transfers complementary knowledge through annealing embedding distillation; and Stage-4 enables efficient deployment with elastic embedding dimensions.


\subsection{Stage-1: Low-Rank Pretraining}
\label{sec:objective}

In Stage-1, we perform contrastive pretraining on the embedding model using the large-scale, omni-modal, and multi-task training corpus. Throughout this stage we use \emph{in-batch mixing}, where training instances from all supported modalities and task types are mixed within each global batch and the candidates of each query are gathered across all data-parallel ranks, so that the negatives of every query span modality and task boundaries.

Concretely, each training instance takes the form of a tuple $(x_i, y_i^{+}, \{y_{i,k}^{-}\}_{k=1}^{K})$ consisting of a query, its positive target, and $K$ accompanying hard negatives. Given a mini-batch of $N$ such instances, all in-batch targets and all in-batch hard negatives are pooled into a single candidate set that is shared by every query, i.e.\ $\mathcal{C} = \{y_j^{+}\}_{j=1}^{N} \cup \{y_{j,k}^{-}\}_{j=1,\,k=1}^{N,\,K}$ of size $N(1+K)$. Every (query, candidate) pair is scored with the cosine similarity
\begin{equation}
\label{eq:cosine}
\mathrm{sim}(x, y) \;=\; \frac{\mathbf{e}(x)^{\top}\mathbf{e}(y)}{\lVert \mathbf{e}(x) \rVert_2 \, \lVert \mathbf{e}(y) \rVert_2} ,
\end{equation}
under a temperature $\tau$, the positive probability and the per-query InfoNCE loss~\citep{infonce} of query $x_i$ are
\begin{equation}
\label{eq:infonce}
\pi_i \;=\; \frac{e^{\mathrm{sim}(x_i, y_i^{+})/\tau}}{Z_i} ,
\qquad
\ell_i \;=\; -\log \pi_i ,
\end{equation}
where the partition function $Z_i$ sums over the whole shared candidate set.
Rather than relying solely on a uniformly averaged InfoNCE objective, our Stage-1 training objective combines two complementary losses: a \emph{focal-weighted contrastive loss} and an \emph{embedding distillation loss}. 

The focal-weighted contrastive loss dynamically reweights queries according to their current retrieval difficulty, reducing the contribution of already well-separated examples and concentrating the optimization budget on unresolved queries with competitive negatives (Section~\ref{sec:focal-loss}). In parallel, the embedding distillation loss transfers the teacher's full similarity distribution over the candidate set, providing graded supervision for both positive and negative candidates instead of the effectively one-hot target used by standard InfoNCE~\citep{hinton_distillation}. This enables the student to preserve fine-grained relevance relationships and ranking structure that cannot be captured by hard labels alone (Section~\ref{sec:distribution-loss}). Together, the two losses provide complementary supervision: the former determines \emph{which queries} should receive greater emphasis, while the latter specifies \emph{how their candidates} should be organized in the embedding space. We introduce the two components in detail below.

\begin{figure*}[t]
\centering
\includegraphics[width=1.0\textwidth]{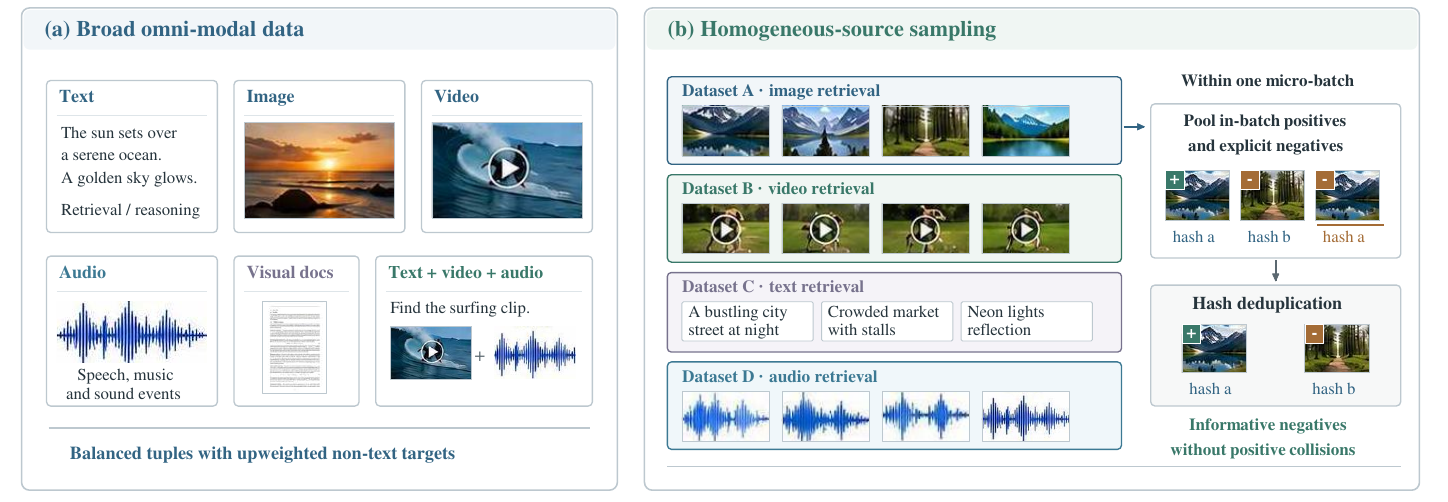}
\caption{\textbf{Data-centric omni-modal training.} (a) Our training corpus
covers text, images, video, audio, visual documents, and interleaved inputs,
with non-text targets upweighted for modality balance. (b) Homogeneous-source
sampling draws each micro-batch from one dataset and deduplicates the pooled
candidates, yielding task-consistent in-batch negatives without positive
collisions.}
\label{fig:method-1}
\end{figure*}

\subsubsection{Focal-Weighted Contrastive Loss}

\label{sec:focal-loss}


Under a uniformly averaged InfoNCE objective, a query whose positive is already well separated from its negatives occupies the same share of the batch loss as a query whose hard negatives remain competitive. Inspired by focal learning for dense object detection~\citep{lin2017focal} and discovery in our recent embedding research~\citep{pace}, the \emph{focal embedding loss} rescales the per-query loss by its current retrieval difficulty, so that queries that are not yet resolved dominate each update.

We use the positive probability $\pi_i$ defined in
Eq.~\eqref{eq:infonce} as the retrieval confidence,
with the same candidate set $\mathcal{C}$ and
temperature $\tau$. The confidence is then mapped to a per-query focal coefficient that is treated as a constant, and the coefficients are normalized to unit mean over the mini-batch:
\begin{equation}
\label{eq:focal-coef}
a_i \;=\; \mathrm{sg}\!\left[\bigl(1 - \pi_i\bigr)^{\gamma}\right],
\qquad
\tilde{w}_i \;=\; \frac{a_i}{\frac{1}{N}\sum_{k=1}^{N} a_k} ,
\end{equation}
where $\mathrm{sg}[\cdot]$ denotes the stop-gradient operator and $\gamma \geq 0$ controls the strength of difficulty modulation, with $\gamma = 0$ recovering uniform weighting. The focal embedding loss is the resulting reweighted contrastive objective
\begin{equation}
\label{eq:focal-loss}
\mathcal{L}_{\mathrm{focal}}
\;=\;
\frac{1}{N}\sum_{i=1}^{N} \tilde{w}_i \, \ell_i
\;=\;
-\,\frac{\sum_{i=1}^{N} a_i \log \pi_i}{\sum_{i=1}^{N} a_i} .
\end{equation}

Queries whose positives are angularly well separated from the competing candidates yield high confidence $\pi_i$ and are correspondingly downweighted, whereas ambiguous queries with competitive negatives retain large weights. Since the unit-mean normalization keeps the overall loss scale unchanged relative to uniform averaging, the reweighting redistributes rather than inflates the optimization budget across the mini-batch.

\subsubsection{Embedding Distillation Loss}
\label{sec:distribution-loss}

The standard contrastive objective treats the positive target as a one-hot label and does not explicitly model the relative relevance of the remaining candidates. To provide richer supervision during embedding model initialization, we introduce a distribution loss that transfers the teacher's full ranking distribution over the candidate set. For each query \(x_i\), the teacher distribution is:
\begin{equation}
\label{eq:teacher-distribution}
t_i(c)=\frac{e^{\mathrm{sim}_{\mathrm{T}}(x_i,c)/\tau}}{\displaystyle\sum_{c'\in\mathcal{C}}e^{\mathrm{sim}_{\mathrm{T}}(x_i,c')/\tau}},\qquad c\in\mathcal{C}
\end{equation}
where \(\mathrm{sim}_{\mathrm{T}}\) denotes the similarity produced by the teacher embedding model. In practice, the candidate-level log-probabilities \(\log t_i(c)\) are computed offline and stored with each training instance, avoiding teacher-model inference during student training.

Using the similarity function defined in Eq.~\eqref{eq:cosine}, the corresponding student distribution is:
\begin{equation}
\label{eq:student-distribution}
q_i(c)=\frac{e^{\mathrm{sim}(x_i,c)/\tau}}{\displaystyle\sum_{c'\in\mathcal{C}}e^{\mathrm{sim}(x_i,c')/\tau}},\qquad c\in\mathcal{C}
\end{equation}
The teacher and student distributions are constructed with the same candidate ordering and temperature. We then minimize the forward Kullback--Leibler divergence from the teacher distribution to the student distribution:
\begin{equation}
\label{eq:distribution-loss}
\begin{aligned}
\ell_i^{\mathrm{KD}}
&= \operatorname{KL}\!\left(t_i\,\middle\|\,q_i\right)
= \sum_{c\in\mathcal{C}}t_i(c)\log\frac{t_i(c)}{q_i(c)},\\
\mathcal{L}_{\mathrm{dist}}
&= \frac{1}{N}\sum_{i=1}^{N}\ell_i^{\mathrm{KD}}.
\end{aligned}
\end{equation}

Unlike one-hot supervision, the teacher distribution assigns graded probability mass to both the positive and negative candidates, thereby preserving their relative relevance and difficulty. Minimizing \(\mathcal{L}_{\mathrm{dist}}\) encourages the student to reproduce the teacher's fine-grained ranking structure rather than merely separating the positive from all negatives. Teacher models may differ across modalities, while their output distributions share the same representation and are optimized through the unified distribution loss above.

\subsubsection{Optimization Objective}
The final Stage-1 objective is a weighted combination of the two terms above,
\begin{equation}
\label{eq:stage1-loss}
\mathcal{L}_{\text{stage-1}} \;=\; \lambda_{\mathrm{focal}} \, \mathcal{L}_{\mathrm{focal}} \;+\; \lambda_{\mathrm{dist}} \, \mathcal{L}_{\mathrm{dist}} ,
\end{equation}
where $\mathcal{L}_{\mathrm{focal}}$ (Eq.~\eqref{eq:focal-loss}) is the focal-weighted contrastive loss, $\mathcal{L}_{\mathrm{dist}}$ is the distribution loss distilled from teacher soft labels, and $\lambda_{\mathrm{focal}}, \lambda_{\mathrm{dist}} \geq 0$ balance hard-negative-focused contrastive learning against teacher-guided distribution matching. Both terms are computed over the same candidate set $\mathcal{C}$. In all our Stage-1 runs we set $\lambda_{\mathrm{focal}} = \lambda_{\mathrm{dist}}$ without further tuning.

\textbf{LoRA Protects Early Embeddings:} In our training, we first observed that full-parameter updates perform markedly worse than low-rank ones. The reason lies in the initial state of the embeddings: when a base model is first adapted, the new representations carry no meaningful constraints—they sit in a chaotic, arbitrarily organized space. Enabling full-parameter training at this point lets unstable gradients move every weight, causing violent jumps that erode the pretrained semantic structure before the embeddings ever stabilize. LoRA avoids this by confining updates to a low-dimensional subspace, limiting their magnitude and acting as a stabilizing prior. This shields the base model's knowledge while the embedding space gradually organizes itself—preserving the semantics that make the model useful until the new representations become coherent.

\subsection{Stage-2: Full-Parameter Homogeneous Finetuning}

After Stage-1 pretraining, the model has acquired basic embedding capabilities and a shared representation space across all modalities. In this stage, we further finetune it on a smaller collection of higher-quality downstream datasets.

We continue to use the InfoNCE loss with in-batch negatives (Eq.~\eqref{eq:infonce}), but change the sampling strategy. Unlike Stage-1, where modalities and datasets are thoroughly mixed, Stage-2 restricts all samples within each micro-batch to the same dataset. Accordingly, the focus of training shifts from establishing a unified representation space to refining the model's representations and fine-grained discrimination. We refer to this strategy as \emph{homogeneous-source sampling}, abbreviated as \emph{homogeneous sampling}.

Formally, let $B_{\mu}$ denote the micro-batch size and $s_i$ the source dataset of the $i$-th instance. A homogeneous micro-batch from dataset $d$, its deduplicated candidate set, and the resulting batch loss are written as
\begin{equation}
\label{eq:homo-sampling}
\begin{aligned}
\mathcal{B}_d
&= \left\{(x_i,y_i^{+},\{y_{i,k}^{-}\}_{k=1}^{K})\right\}_{i=1}^{B_{\mu}},
\,\,\, s_i=d\ \;\forall i, \\
\mathcal{C}(\mathcal{B}_d)
&= \operatorname{Dedup}_{\mathrm{hash}}\!\left(
\{y_j^{+}\}_{j=1}^{B_{\mu}} \cup
\{y_{j,k}^{-}\}_{j=1,\,k=1}^{B_{\mu},\,K}\right), \\
\mathcal{L}(\mathcal{B}_d)
&= -\frac{1}{B_{\mu}}\sum_{i=1}^{B_{\mu}}\log
\frac{e^{\mathrm{sim}(x_i,y_i^{+})/\tau}}
{\sum_{y\in\mathcal{C}(\mathcal{B}_d)}e^{\mathrm{sim}(x_i,y)/\tau}} .
\end{aligned}
\end{equation}


When in-batch negatives are enabled, the candidates associated with other instances in the same micro-batch---including their positive and negative candidates---are gathered and used as additional negatives for the current query. In a randomly mixed batch, these instances often come from different modalities or task types. The model may therefore learn shortcuts based on superficial cues such as modality, sentence length, or linguistic style, rather than making fine-grained semantic distinctions. Homogeneous sampling addresses this issue by drawing every instance in a micro-batch from the same dataset, making the gathered negatives more informative and providing stronger supervision for fine-grained discrimination (c.f. Fig.~\ref{fig:method-1}).

In implementation, we deduplicate candidates by hash. This prevents a gathered negative from being identical to the positive candidate, which would create a false negative, and removes duplicates of the explicit negative candidates. Homogeneity is enforced only within each micro-batch, so an optimization step can still average gradients from different datasets. This reduces update oscillation and mitigates catastrophic forgetting. As a result, homogeneous sampling can be applied beyond dense retrieval to tasks such as classification and clustering, where it also yields consistent improvements.

Homogeneous sampling also substantially increases the proportion of regularly shaped micro-batches, since instances from the same dataset usually contain the same number of candidates. This property brings considerable training speedups in frameworks such as MS-SWIFT~\citep{msswift}. Compared with random sampling, the strategy improves performance  across benchmarks, and the gain grows further as the micro-batch size increases.


\begin{figure*}[t]
\centering
\includegraphics[width=1.0\textwidth]{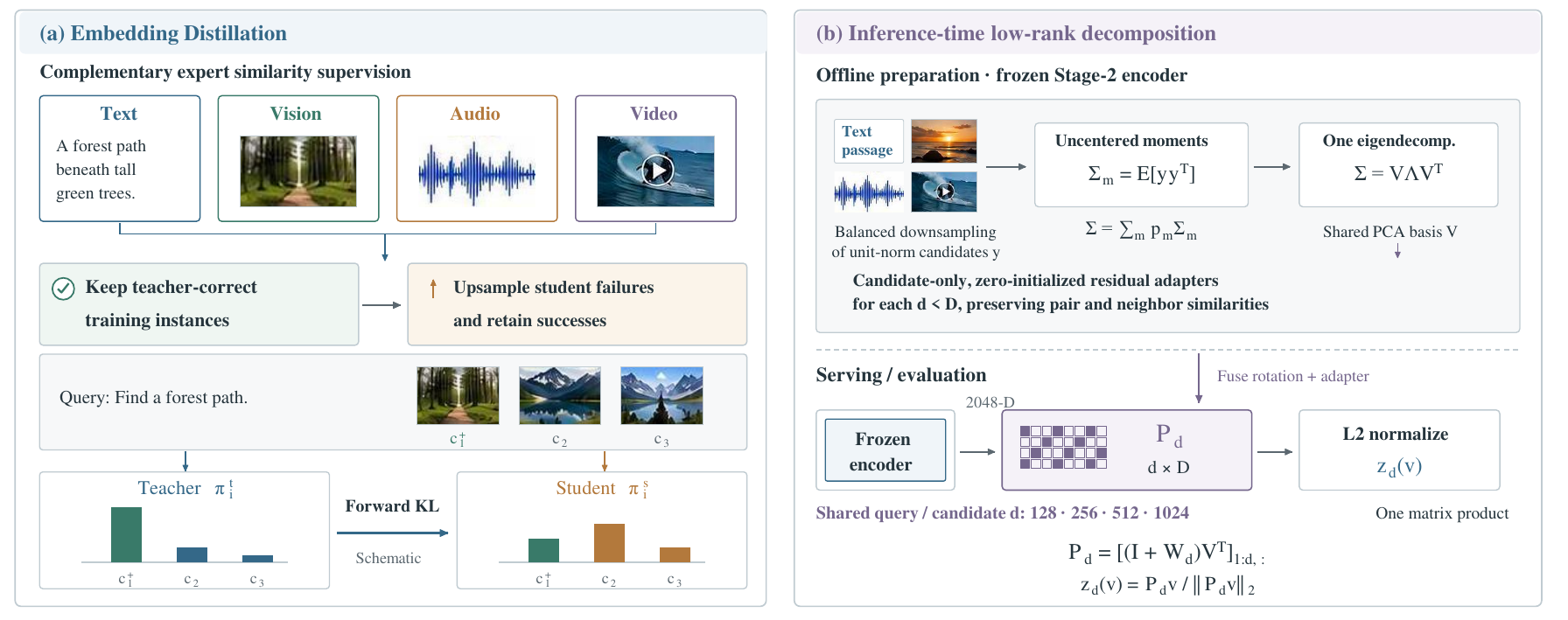}
\caption{\textbf{Embedding-specific training and inference optimization.}
(a) Embedding Distillation transfers similarity distributions from
complementary modality experts, retaining teacher-correct examples and
emphasizing student failures through forward KL supervision. (b) A shared PCA
basis and lightweight residual adapters produce compact embeddings at multiple
dimensions; the rotation and adapter are fused into a single projection for
efficient inference.}
\label{fig:method-2}
\end{figure*}

\subsection{Stage-3: Annealing Embedding Distillation}

\label{sec:opd}

After Stage-2 finetuning, the model's performance begins to saturate. Limited by the capabilities of the base model and the student's capacity, continued training on the same data yields little further improvement. To move beyond this limit, we introduce \emph{Embedding Distillation} (ED), which transfers knowledge from a stronger teacher to the student (see Fig.~\ref{fig:method-2}).

We first run the teacher on the same training corpus used in Stage-2 and retain only the instances that it solves correctly. We then evaluate the student on the retained instances and divide them according to whether the student succeeds or fails. Student failures are upsampled to form the ED training set, directing more of the training budget toward capabilities for which useful knowledge can be transferred from the teacher.

ED retains both in-batch negatives and homogeneous sampling, ensuring that the student receives a sufficiently large set of negatives of moderate difficulty. Because the in-batch candidate pool is formed randomly at each step, its negative distribution continually changes. These additional negatives drawn from the same source allow the student to distill the teacher's distribution more thoroughly. For each query $x_i$, we use the teacher distribution $t_i(c)$
and student distribution $q_i(c)$ defined in
Equations~\ref{eq:teacher-distribution}
and~\ref{eq:student-distribution}, evaluated with the teacher,
student, and candidate set $\mathcal{C}$ of this stage. The per-query distillation loss $\ell_i^{\mathrm{KD}}$ is
the forward KL divergence defined in
Eq.~\eqref{eq:distribution-loss}, evaluated using
the distributions of this stage.
We use forward rather than reverse KL because the main InfoNCE objective already concentrates probability on the positive candidate, while reverse KL is also mode-seeking and therefore partly duplicates this effect. Forward KL instead transfers the teacher's softer ranking over the full candidate set, providing complementary supervision among both positive and negative candidates.

Our experiments reveal a useful trade-off. Student-only InfoNCE training, which resembles annealed training with upsampled hard examples, further improves tasks on which the student is already strong. In contrast, assigning a large fixed weight to KL substantially improves tasks where the student is weak but the teacher is strong. We therefore balance learning across domains with an adaptive, per-query distillation weight determined by the student's confidence in the positive candidate.

Using the InfoNCE loss $\ell_i$ defined in
Eq.~\eqref{eq:infonce}, with $\pi_i=q_i(y_i^{+})$,
we define the query difficulty as
\begin{equation}
\label{eq:opd-difficulty}
d_i
=
\left(1-e^{-\ell_i}\right)^{\gamma}
=
\left(1-\pi_i\right)^{\gamma}.
\end{equation}
where $\gamma>0$ controls how quickly the teacher's influence decreases as the student becomes more confident. Since $d_i \in [0,1]$, we obtain a bounded distillation weight
\begin{equation}
\label{eq:opd-weight}
\lambda_i
=
\lambda_{\min}
+
\left(\lambda_{\max}-\lambda_{\min}\right)
\mathrm{sg}\!\left[d_i\right],
\end{equation}
where $0 \leq \lambda_{\min} \leq \lambda_{\max}$ and
$\mathrm{sg}[\cdot]$ is the stop-gradient operator.
The final ED objective is
\begin{equation}
\label{eq:opd-objective}
\mathcal{L}_{\mathrm{ED}}
=
\underbrace{\frac{1}{N}\sum_{i=1}^{N}\ell_i}
_{\mathcal{L}_{\mathrm{NCE}}^{\mathrm{s}}}
+
\frac{1}{N}\sum_{i=1}^{N}\lambda_i \ell_i^{\mathrm{KD}} .
\end{equation}
The KL weights are averaged over the fixed batch size $N$, rather than normalized by their sum, so the overall influence of the teacher can decrease as the student masters more training instances. Confident queries receive weights near $\lambda_{\min}$, whereas uncertain queries receive stronger teacher supervision approaching $\lambda_{\max}$. ED further improves performance across omni-modal benchmarks. At the task level, it strengthens areas in which the student is already competitive, such as classification and question answering, while also improving previously weaker capabilities such as retrieval.

\subsection{Stage-4: Elastic Embedding Inference}
\label{sec:exp-analysis}

Having established a strong shared embedding space through the preceding training stages, we finally turn to adapting the resulting representation for efficient and flexible deployment.

Deployed retrieval systems face heterogeneous storage, latency and accuracy budgets, yet training a separate encoder for each configuration is prohibitively expensive. Although Matryoshka Representation Learning (MRL)~\citep{mrl} enables a single encoder to support multiple embedding widths, we observe that incorporating it into multi-objective training can degrade the performance of full-dimensional embeddings. To decouple multi-width adaptation from encoder training, we therefore attach a lightweight adapter after stage-2 training. This adapter makes a single 2,048-dimensional embedding usable at multiple nested widths while preserving the original embedding’s semantic similarity, nearest-neighbor structure and ranking order. Please refer to Fig.~\ref{fig:method-2} for details.

\paragraph{Task formulation.}
Let $E$ denote the encoder frozen after above training, producing unit-norm
query and candidate embeddings $\mathbf{x}, \mathbf{y} \in \mathbb{R}^{D}$ with $D = 2048$,
and let $\mathcal{D} = \{128, 256, 512, 1024, 2048\}$ be the nested width set. For
each $d \in \mathcal{D}$ with $d < D$ we learn an adapter
$f_d : \mathbb{R}^{D} \to \mathbb{R}^{D}$ and define the width-$d$
representation as its renormalized leading prefix,
\begin{equation}
  z^{(d)}(v) \;=\; \frac{u_{1:d}}{\bigl\lVert u_{1:d} \bigr\rVert_2}
  \quad \text{with} \quad u = f_d(V^{\!\top} v),
  \qquad
  \operatorname{sim}_d(\mathbf{x}, \mathbf{y})
  \;=\;
  \bigl\langle z^{(d)}(\mathbf{x}),\, z^{(d)}(\mathbf{y}) \bigr\rangle ,
  \label{eq:readout}
\end{equation}
where $V$ is the shared orthogonal transform of Eq.~\eqref{eq:hybridpca} below.
Renormalization is required because truncating a unit-norm
vector shrinks its norm, which otherwise breaks the agreement between
cosine, inner product and $\ell_2$ distance.

\paragraph{Implementation Details.}
Each adapter is preceded by a shared, training-free orthogonal transform
and instantiated as a zero-initialized residual linear map,
\begin{equation}
  f_d(\tilde{v}) \;=\; \tilde{v} + W_d \tilde{v},
  \qquad W_d \big|_{\mathrm{init}} = 0,
  \qquad \tilde{v} = V^{\!\top} v .
  \label{eq:pipeline}
\end{equation}

\emph{Hybrid PCA transform.} Write $\mathcal{C}_1, \dots, \mathcal{C}_M$ for
the per-modality candidate pools. For each we estimate the uncentered
second moment by balanced downsampling, and mix the moments under a
weight vector $p = (p_1, \dots, p_M)$ with $\sum_m p_m = 1$:
\begin{equation}
  \Sigma \;=\; \sum_{m} p_m \,
          \mathbb{E}_{\mathbf{y} \sim \mathcal{C}_m}
          \!\left[\, \mathbf{y}\, \mathbf{y}^{\!\top} \right],
  \qquad \Sigma = V \Lambda V^{\!\top},
  \qquad \mu_1 \ge \cdots \ge \mu_D .
  \label{eq:hybridpca}
\end{equation}
Here $\Lambda = \operatorname{diag}(\mu_1,\dots,\mu_D)$.

 Since $V$ is orthogonal, $V^{\!\top} v$ is an isometry of $\mathbb{R}^{D}$
--- a rigid re-expression of the embedding that leaves all inner
products, and hence the full-width geometry exactly unchanged.

PCA is an established post-hoc compressor for text
retrieval~\citep{pca,tongj}, but once the corpus is multimodal it is
no longer obvious \emph{which} embeddings the basis should be fitted on. A
preliminary probe comparing per-modality PCA subspaces shows that they
agree on only a handful of leading directions and diverge rapidly beyond
them: prefix capacity is contested, and $p$ is what allocates it. Because
the corpora differ in size by orders of magnitude, each pool is
downsampled to a common budget; otherwise sample counts, not $p$, would
determine the basis. The mixture is formed over second moments, with a
single eigendecomposition taken at the end---averaging per-modality bases
would instead yield a non-orthogonal operator and forfeit the invariance
established above. Ablating $p$, we find uniform weights to be a robust
default.

\emph{Residual adapter.} The PCA basis alone already improves the short
widths, but by less than we had expected, so we additionally train a
residual adapter in the rotated space.
Training is unsupervised: $f_d$ is fitted on the candidate corpus $\{\mathbf{y}\}$
alone, queries enter only at evaluation time, and the objective below acts
as a proxy for how far the width-$d$ prefix geometry has drifted from the full-width. Each
step forms its batch from a single dataset, drawn under a three-level
uniform schedule over modality $\to$ category $\to$ dataset so that every
modality receives an equal share of steps.

We adopt the unsupervised objective of the Matryoshka-Adaptor
\citep{matryoshka}, which combines a global and a local
similarity-preservation term over a batch $B$ against the adapter-free
target $\operatorname{sim}_D$:
\begin{align}
  \mathcal{L}^{\mathrm{pair}}_d
  &= \frac{1}{|B|\,(|B|-1)} \sum_{i \in B} \sum_{j \in B,\, j \neq i}
     \Bigl\lvert\, \operatorname{sim}_d(\mathbf{y}_i, \mathbf{y}_j)
                 - \operatorname{sim}_D(\mathbf{y}_i, \mathbf{y}_j) \,\Bigr\rvert ,
  \label{eq:lpair}\\
  \mathcal{L}^{\mathrm{topk}}_d
  &= \frac{1}{|B|\,S} \sum_{i \in B} \sum_{k=1}^{S}
     \Bigl\lvert\, \operatorname{sim}_d\bigl(\mathbf{y}_i, \mathbf{y}_{n_k(i)}\bigr)
                 - \operatorname{sim}_D\bigl(\mathbf{y}_i, \mathbf{y}_{n_k(i)}\bigr) \,\Bigr\rvert ,
  \label{eq:ltopk}\\
  \mathcal{L}_d
  &= \mathcal{L}^{\mathrm{topk}}_d + \alpha\, \mathcal{L}^{\mathrm{pair}}_d .
  \label{eq:total}
\end{align}
Eq.~\eqref{eq:lpair} compares every pair within a batch against the
full-width Gram matrix; as most such pairs are unrelated, it mainly pins
down the coarse arrangement of the corpus. Eq.~\eqref{eq:ltopk} looks only
at each anchor and $S$ of its precomputed nearest neighbors, the region
where retrieval metrics are actually decided. Both targets are detached,
and deviations are measured in $\ell_1$ by default. Instead of adopting a shallow MLP like Matryoshka-Adaptor, we find that
once the skip connection is in place the choice of shallow structure makes
little difference, and take the single linear residual map of
Eq.~\eqref{eq:pipeline} as the most economical of the structures we tried. Moreover, linearity lets the
width-$d$ embedding fold into a single $d \times D$ matrix,
$(I + W_d) V^{\!\top}$ truncated to its first $d$ rows, so serving one
width costs one matrix product.


\begin{figure}[t]
\centering
\begin{subfigure}[t]{0.48\linewidth}
  \centering
  \includegraphics[width=\linewidth]{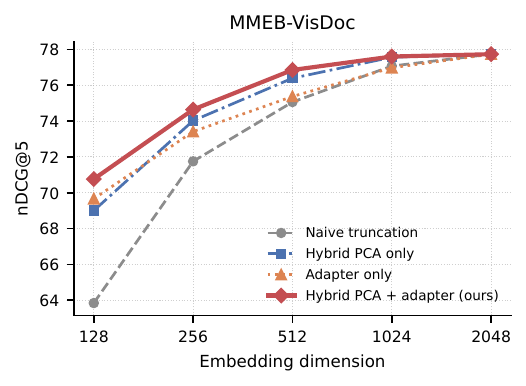}
  \caption{MMEB-VisDoc (nDCG@5)}
  \label{fig:curve-visdoc}
\end{subfigure}\hfill
\begin{subfigure}[t]{0.48\linewidth}
  \centering
  \includegraphics[width=\linewidth]{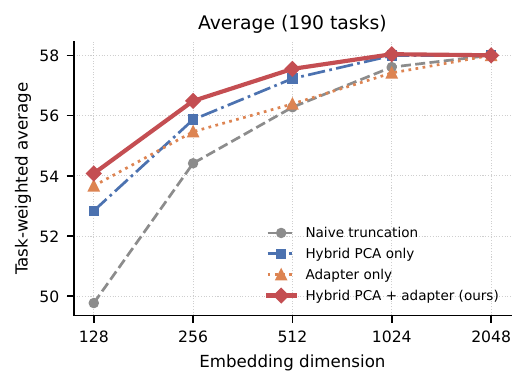}
  \caption{MMEB-Average}
  \label{fig:curve-avg}
\end{subfigure}
\caption{\textbf{Performance versus embedding width.} We compare naive truncation of
  the frozen encoder, the hybrid PCA basis alone, the residual adapter alone, and the
  two combined; (b) averages the six suites with the task weights of
  Table~\ref{tab:elastic-dims}. The components are complementary: the PCA basis carries
  most of the gain
  at $d\ge512$, whereas the adapter dominates at $d=128$; neither alone matches the
  combination.}
\label{fig:curves}
\end{figure}

\section{Evaluation}
\label{sec:exp}
In this section we empirically evaluate Ovis-Embedding across a comprehensive suite of multimodal retrieval benchmarks, with the goal of answering three questions: (i)~does a single omni-modal encoder match or surpass strong modality-specialist baselines on each individual modality pair; (ii)~which components of the proposed recipe contribute the most to its quality; and (iii)~how does the learned representation behave qualitatively on out-of-distribution queries. We first describe the evaluation setup and baselines (Section~\ref{sec:exp-setup}), then report the main quantitative results (Section~\ref{sec:exp-main}).

\subsection{Experimental Setup}
\label{sec:exp-setup}

\paragraph{Benchmarks.}
We evaluate Ovis-Embedding on five complementary benchmark families that
collectively test vision--language, omni-modal, audio, video, and
retrieval-specific representations:
\begin{itemize}\setlength{\itemsep}{1pt}
  \item \textbf{MMEB-v3}~\citep{mmebv3} is our primary omni-modal benchmark. It extends
    MMEB-v2 to 190 tasks spanning text, images, video, audio, visual
    documents, and agent-centric scenarios. In addition to conventional
    classification, question answering, retrieval, and grounding, it includes
    complex text retrieval, cross-modal retrieval involving audio, and tool,
    GUI, and memory retrieval. We use it to assess whether
    Ovis-Embedding-Omni-3B provides balanced performance across all supported
    modalities and application settings.

  \item \textbf{MMEB-v2}~\citep{vlm2vec_v2} evaluates vision--language embeddings over images,
    video, and visual documents. Its image tasks cover classification, visual
    question answering, image retrieval, and visual grounding; its video tasks
    cover classification, question answering, retrieval, and temporal
    grounding; and its visual-document tasks evaluate page- and document-level
    retrieval. We use MMEB-v2 as the principal benchmark for
    Ovis-Embedding-VL-2B and Ovis-Embedding-VL-9B.

  \item \textbf{MAEB} (Massive Audio Embedding Benchmark)~\citep{maeb} contains 30 tasks
    covering speech, music, environmental sounds, bioacoustics, and
    multilingual audio. It evaluates both acoustic and linguistic information
    through classification, clustering, pair classification, reranking, audio
    retrieval, and audio--text alignment.

  \item \textbf{MVEB} (Massive Video Embedding Benchmark)~\citep{mveb} contains 23 tasks
    spanning video classification, zero-shot classification, clustering, pair
    classification, retrieval, and video-centric question answering. It
    includes both video-only and audio--video settings, allowing us to measure
    temporal visual understanding as well as the contribution of the
    accompanying audio stream.

  \item \textbf{RTEB} (Retrieval Embedding Benchmark)~\citep{rteb} focuses on text
    retrieval for realistic search and retrieval-augmented generation. It
    combines open and private evaluation sets across domains such as law,
    finance, programming, and healthcare, and therefore complements the broad
    multimodal suites with a dedicated test of retrieval quality and
    out-of-distribution generalization.
\end{itemize}

We report the official aggregate score for each benchmark rather than
constructing a new average across benchmark families. For MMEB-v2 and MMEB-v3,
the overall score is the unweighted average over their constituent datasets;
we additionally report modality- and task-level breakdowns. MAEB, MVEB, and
RTEB are aggregated using their official evaluation implementations.

\paragraph{Evaluation protocol.}
All inputs are encoded independently in a bi-encoder setting. For
Ovis-Embedding, each query is paired with the task instruction defined by the
corresponding benchmark and formatted using the backbone's native processor
and chat template. We extract the final-layer hidden state at the last
non-padding token and normalize it to unit length. Candidates are encoded
independently in the same way, and rankings are produced by cosine similarity.
No answer generation or cross-encoder interaction is used during evaluation:
classification labels, candidate answers, passages, and multimodal items are
all treated as retrieval candidates in the shared embedding space.

We evaluate baselines using their officially released checkpoints,
preprocessing pipelines, prompts, and pooling rules, avoiding model-specific
disadvantages from imposing a single extraction strategy on architectures with
different designs. Each dataset is scored with the metric prescribed by its
benchmark. MMEB primarily uses Hit@1 for image, video, audio, and agent tasks
and nDCG@5 for text and visual-document retrieval, while RTEB uses nDCG@10 as
its default leaderboard metric. MAEB and MVEB use the task-specific metrics
implemented in MTEB.
Unless otherwise stated, all reported values are multiplied by 100, and the
same candidate pools and benchmark splits are used for every compared model.

\paragraph{Implementation details.}
We evaluate three Ovis-Embedding variants.
\textbf{Ovis-Embedding-Omni-3B} is initialized from Qwen2.5-Omni-3B and
supports text, images, video, audio, and interleaved combinations of these
modalities. \textbf{Ovis-Embedding-VL-2B} and
\textbf{Ovis-Embedding-VL-9B} are initialized from Qwen3.5-2B and Qwen3.5-9B,
respectively, and support text, images, video, and their interleaved
combinations. Because no embedding projection head is introduced, the output
dimension is inherited from each backbone: 2,048 for Omni-3B and VL-2B, and
4,096 for VL-9B.

The models are trained with the three training stages described in
Section~\ref{sec:training}: low-rank pretraining, full-parameter homogeneous finetuning, and annealing embedding distillation. All experiments are
conducted on clusters equipped with NVIDIA H100 80\,GB GPUs. During evaluation,
maximum sequence lengths, image resolution, video frame sampling, audio
preprocessing, and other modality-specific settings follow the official
benchmark configuration and are fixed across models whenever their native
processors permit.

\subsection{Omni-3B Results}
\label{sec:exp-main}

We evaluate \textbf{Ovis-Embedding-Omni-3B} on three complementary
omni-modal benchmark suites. MMEB-v3 measures general-purpose embedding
quality across image, video, visual-document, text, audio, and agent tasks,
whereas MAEB and MVEB provide more focused evaluations of audio and
video representations, respectively. We also evaluate our Omni-3B model on broader text benchmarks (e.g. RTEB) to show our generalization ability. Together, these benchmarks
test not only whether a single 3B model can cover heterogeneous modalities,
but also whether it retains the modality-specific discrimination required by
classification, retrieval, clustering, reranking, and question answering. All reported scores are percentages, and higher is
better.

\subsubsection{MMEB-v3}

Table~\ref{tab:mmeb-v3} compares
\textbf{Ovis-Embedding-Omni-3B} with representative omni-modal embedding models on MMEB-v3. The
benchmark covers six evaluation groups---image, video, visual document, text,
audio, and agent---and therefore provides a direct test of whether a single
compact model can maintain strong performance across heterogeneous modalities
and retrieval settings.

\paragraph{Overall performance.}
Ovis-Embedding-Omni-3B achieves an overall score of \textbf{58.46}, exceeding
the strongest baseline, Tianmu-Emb-Uni~\citep{tianmu_emb_uni} (53.27), by 5.19 points. It also
outperforms e5-omni-7B (47.14) and Omni-Embed-Nemotron-3B~\citep{omni_embed_nemotron} (43.60) by 11.32 and
14.86 points, respectively, despite using only 3B parameters. More
importantly, Ovis-Embedding-Omni-3B ranks first on the aggregate score of every
evaluation group: 77.55 on image, 64.99 on video, 78.26 on visual documents,
47.15 on text, 50.08 on audio, and 45.52 on agent tasks. The corresponding
margins over the second-best model are 3.72, 5.62, 2.89, 3.53, 7.04, and 6.10
points. This consistent advantage shows that the overall gain does not arise
from a single dominant modality, but from broadly improved representations
across the full MMEB-v3 task spectrum.

\begin{table*}[t]
  \centering
  \caption{\textbf{Results on MMEB-v3.}
  We compare Ovis-Embedding-Omni-3B with representative omni-modal embedding
  models across six modalities and their constituent sub-tasks. Scores are
  reported as percentages. The best and second-best results in each row are
  shown in \textbf{bold} and \underline{underlined}, respectively.}
  \label{tab:mmeb-v3}
  \small
  \setlength{\tabcolsep}{6pt}
  \renewcommand{\arraystretch}{1.04}
  \renewcommand{\multirowsetup}{\centering}
  \begin{tabular*}{\textwidth}{@{}>{\centering\arraybackslash}m{0.11\textwidth}@{\hspace{4pt}}>{\centering\arraybackslash}m{0.16\textwidth}@{\hspace{4pt}\extracolsep{\fill}}ccccc@{}}
    \toprule
    \multirow[c]{2}{=}{\raisebox{-1.4\baselineskip}{\textbf{Modality}}} &
    \multirow[c]{2}{=}{\raisebox{-1.4\baselineskip}{\textbf{Sub-task}}} &
    \multicolumn{5}{c}{\textbf{Model}} \\
    \cmidrule(r){3-7}
    & &
    {\footnotesize\raisebox{0.5\baselineskip}{\shortstack[c]{\textbf{Ovis-Embedding-}\\\textbf{Omni-3B}}}} &    
    {\footnotesize\shortstack[c]{LCO-\\Embedding-\\Omni-7B}} &
    {\footnotesize\raisebox{0.5\baselineskip}{\shortstack[c]{omni-embed-\\nemotron-3b}}} &
    {\footnotesize\raisebox{0.5\baselineskip}{\shortstack[c]{e5-omni-\\7B}}} &
    {\footnotesize\shortstack[c]{Tianmu-\\Emb-Uni\\(8B)}} \\
    \midrule

    \multirow[c]{5}{=}{\textbf{IMG}}
      & \textit{Overall} & \textbf{77.55} & 56.27 & 43.42 & 70.50 & \underline{73.83} \\
      & I-CLS            & \textbf{76.83} & 57.97 & 48.31 & \underline{68.19} & 65.99 \\
      & I-QA             & \underline{77.70} & 51.83 & 19.90 & 70.30 & \textbf{78.00} \\
      & I-RET            & \textbf{72.86} & 53.88 & 56.06 & 69.65 & \underline{72.35} \\
      & I-VG             & \textbf{94.17} & 70.90 & 48.90 & 79.55 & \underline{87.80} \\
    \midrule

    \multirow[c]{5}{=}{\textbf{VID}}
      & \textit{Overall} & \textbf{64.99} & 53.03 & 41.36 & 50.83 & \underline{59.37} \\
      & V-CLS            & \textbf{80.26} & 58.11 & 48.31 & 53.86 & \underline{66.95} \\
      & V-QA             & \textbf{67.73} & 62.36 & 47.81 & 65.89 & \underline{67.51} \\
      & V-RET            & \textbf{52.05} & 42.23 & 38.68 & 40.50 & \underline{50.21} \\
      & V-MRET           & \textbf{56.57} & 46.99 & 23.50 & 37.88 & \underline{48.43} \\
    \midrule

    \multirow[c]{5}{=}{\textbf{VISDOC}}
      & \textit{Overall} & \textbf{78.26} & 69.19 & 72.27 & \underline{75.37} & 72.03 \\
      & ViDoRe-V1        & \underline{86.67} & 80.08 & 84.76 & \textbf{87.33} & 83.69 \\
      & ViDoRe-V2        & \underline{56.22} & 54.14 & 51.57 & \textbf{58.33} & 53.19 \\
      & VisRAG           & \underline{85.81} & 79.55 & 85.39 & \textbf{87.60} & 83.50 \\
      & VisDoc-OOD       & \textbf{67.94} & 41.47 & 42.04 & 44.19 & \underline{44.49} \\
    \midrule

    \multirow[c]{8}{=}{\textbf{TXT}}
      & \textit{Overall} & \textbf{47.15} & 32.40 & 39.23 & 26.93 & \underline{43.62} \\
      & FollowIR         & \textbf{53.96} & 37.39 & 31.74 & 26.41 & \underline{39.03} \\
      & R2MED            & \underline{25.68} & 12.27 & 17.76 & 18.95 & \textbf{28.09} \\
      & InfoSearch       & \textbf{75.08} & \underline{59.25} & 48.37 & 49.90 & 51.20 \\
      & BRIGHT           & \underline{15.66} & 7.62 & 13.60 & 7.30 & \textbf{15.70} \\
      & LongEmbed        & \textbf{64.00} & 12.57 & 50.03 & 9.71 & \underline{58.49} \\
      & MultiConIR       & 61.73 & 61.36 & \textbf{69.67} & 56.93 & \underline{63.42} \\
      & NanoBEIR         & \underline{61.59} & 52.12 & 56.92 & 35.88 & \textbf{62.05} \\
    \midrule

    \multirow[c]{3}{=}{\textbf{AUD}}
      & \textit{Overall} & \textbf{50.08} & \underline{43.17} & 36.52 & 43.04 & 38.94 \\
      & A-CLS            & \textbf{73.30} & 64.33 & 53.69 & 63.00 & \underline{69.33} \\
      & A-RET            & \textbf{30.73} & 25.53 & 22.21 & \underline{26.41} & 13.62 \\
    \midrule

    \multirow[c]{4}{=}{\textbf{AGENT}}
      & \textit{Overall} & \textbf{45.52} & 27.84 & 36.53 & 36.67 & \underline{39.42} \\
      & Tool             & \textbf{47.56} & 29.05 & 38.05 & 37.45 & \underline{42.21} \\
      & GUI              & \textbf{44.59} & 24.98 & 32.04 & \underline{37.96} & 35.64 \\
      & Memory           & \underline{29.44} & 22.98 & \textbf{32.23} & 27.29 & 22.58 \\
    \midrule

    \textbf{ALL} & \textit{Overall} & \textbf{58.46} & 43.14 & 43.60 & 47.14 & \underline{53.27} \\

    \bottomrule
  \end{tabular*}
\end{table*}

\paragraph{Image and video.}
On image tasks, Ovis-Embedding-Omni-3B obtains the best results on
classification (76.83), retrieval (72.86), and visual grounding (94.17), with
particularly clear gains of 8.64 points on classification and 6.37 points on
grounding over the respective runners-up. Its image-QA score of 77.70 is only
0.30 points below the best result, indicating that the improvement in
retrieval-oriented tasks does not come at the expense of visual question
answering. The advantage is even more consistent for video: our model ranks
first on all four sub-tasks, reaching 80.26 on classification, 67.73 on video
QA, 52.05 on retrieval, and 56.57 on moment retrieval. Compared with the
second-best results, the gains are 13.31 points for video classification and
8.14 points for moment retrieval, demonstrating strong recognition of both
video semantics and temporal structure.

\paragraph{Visual-document and text retrieval.}
Ovis-Embedding-Omni-3B achieves the best visual-document aggregate score of
78.26. Although e5-omni-7B remains slightly stronger on ViDoRe-V1,
ViDoRe-V2, and VisRAG, our model is competitive on these established suites and
substantially improves VisDoc-OOD to 67.94---23.45 points above the second-best
result. This large out-of-distribution gain suggests stronger transfer beyond
the document distributions represented by the standard benchmarks. On text,
our model leads overall with 47.15 and performs particularly well on FollowIR
(53.96), InfoSearch (75.08), and LongEmbed (64.00), outperforming the
corresponding runners-up by 14.93, 23.88, and 5.51 points. It is also close to
the best results on R2MED, BRIGHT, and NanoBEIR. MultiConIR is the main
exception: its score of 61.73 trails Omni-Embed-Nemotron-3B by 7.94 points,
indicating that retrieval under multiple simultaneous constraints remains an
area for further improvement.

\paragraph{Audio and agent tasks.}
The largest modality-level margin appears on audio, where
Ovis-Embedding-Omni-3B reaches 50.08 overall, 7.04 points above the next-best
model. It leads both audio classification (73.30) and audio retrieval (30.73),
improving over the respective runners-up by 3.97 and 4.32 points. These results
show that the model preserves discriminative acoustic information while also
aligning audio with the shared retrieval space. On agent tasks, our model
achieves the best overall score of 45.52, supported by leading results on tool
retrieval (47.56) and GUI retrieval (44.59). It ranks second on memory
retrieval with 29.44, 2.79 points behind Omni-Embed-Nemotron-3B, suggesting
that long-horizon memory matching remains complementary to the model's
strengths in tool and interface understanding.

\paragraph{Summary.}
Across the 31 aggregate and sub-task entries in
Table~\ref{tab:mmeb-v3}, Ovis-Embedding-Omni-3B ranks first on 22 and second on
8; MultiConIR is the only entry on which it falls outside the top two. Taken
together, the results establish Ovis-Embedding-Omni-3B as a strong compact
omni-modal embedder: it combines state-of-the-art overall performance with
balanced modality coverage, while its remaining gaps are concentrated in a
small number of fine-grained document, multi-condition text, and memory
retrieval tasks.


\subsubsection{MAEB \& MVEB}

We further evaluate on the beta versions of the Massive Audio Embedding
Benchmark (MAEB) and the Massive Video Embedding Benchmark
(MVEB). MAEB evaluates audio
representations across 30 tasks covering speech, music, environmental sounds,
and audio--text matching, organized into seven task types. MVEB comprises 23
video tasks spanning zero-shot and supervised classification, retrieval, clustering, pair
classification, and video-centric question answering. These two suites
complement MMEB-v3 by probing whether a unified embedding space preserves
fine-grained acoustic and temporal information. Consistent with the official
leaderboard\footnote{\url{https://huggingface.co/spaces/mteb/leaderboard}}, Tables~\ref{tab:maeb-hf-top10} and
\ref{tab:mveb-hf-top10} report both the average over individual tasks
(Mean(Task)) and the average over task-type aggregates (Mean(Type)). Since
Ovis-Embedding-Omni-3B has not yet been submitted to the official leaderboard,
the displayed ranks are estimated by inserting our local results into the
leaderboard snapshot and recomputing its task-level Borda ranking.


\begin{table*}[t]
  \centering
  \caption{\textbf{Comparison with the top-ranked models on MAEB(beta).}
  Scores are percentages and are aggregated according to the official task types.
  Rows are ordered by Mean(Task) from low to high. ``Rank'' denotes the
  Borda rank after inserting the locally evaluated
  \textbf{Ovis-Embedding-Omni-3B} results. Best results among the models shown are in \textbf{bold}. PC = Audio Pair Classification; M.Clf = Audio Multilabel Classification;
      Retr = Any-to-Any Retrieval; AC = Audio Classification; ZS-Clf = Audio Zero-shot Classification; Clust = Audio Clustering;
      Rerank = Audio Reranking.} 
  \label{tab:maeb-hf-top10}
  \small
  \setlength{\tabcolsep}{3.2pt}
  \renewcommand{\arraystretch}{1.22}
  \resizebox{\textwidth}{!}{%
  \begin{tabular}{lcccccccccc}
    \toprule
    \multicolumn{1}{c}{\multirow{2}{*}{\textbf{Model}}} &
      & \multicolumn{2}{c}{\textbf{Overall}} &
      \multicolumn{7}{c}{\textbf{TaskType Aggregation}} \\
    \cmidrule(lr){3-4}\cmidrule(l){5-11}
    &
    \textbf{Rank} &
    \textbf{Mean(Task)} & \textbf{Mean(Type)} &
    \textbf{PC} & \textbf{M.Clf} & \textbf{Retr} & \textbf{AC} &
    \textbf{ZS-Clf} & \textbf{Clust} & \textbf{Rerank} \\
    \midrule
    Qwen2-Audio-7B & 7 & 34.54 & 37.01 &
      56.89 & 32.64 & 5.29 & 58.34 & 12.38 & 12.65 & 80.85 \\
    larger\_clap\_general & 8 & 34.98 & 41.08 &
      51.89 & \textbf{80.63} & 18.91 & 47.88 & 14.87 & 6.62 & 66.78 \\
    OmniEmbed-v0.1 & 4 & 43.87 & 47.90 &
      58.90 & 68.87 & 41.95 & 49.06 & 30.16 & 3.22 & 83.14 \\
    e5-omni-3B & 11 & 48.00 & 51.60 &
      59.48 & 51.54 & 51.21 & 48.19 & 64.43 & 1.86 & 84.47 \\
    jina-embeddings-v5-omni-nano & 10 & 50.14 & 55.31 &
      65.56 & 70.83 & 50.03 & 51.34 & 60.26 & 5.96 & 83.19 \\
    jina-embeddings-v5-omni-small & 5 & 50.41 & 55.58 &
      65.38 & 69.26 & 49.76 & 51.99 & 62.16 & 6.13 & 84.38 \\
    LCO-Embedding-Omni-3B & 9 & 52.25 & 55.50 &
      66.66 & 74.05 & 54.55 & 54.46 & 62.17 & 1.27 & 75.37 \\
    BidirLM-Omni-2.5B-Embedding & 6 & 52.36 & 54.52 &
      66.75 & 60.95 & 47.47 & \textbf{59.77} & 65.77 & 6.18 & 74.76 \\
    e5-omni-7B & 2 & 52.45 & 57.12 &
      65.54 & 68.78 & 56.07 & 52.17 & \textbf{68.81} & 1.78 & \textbf{86.70} \\
    LCO-Embedding-Omni-7B & 3 & 53.54 & 57.06 &
      67.30 & 75.74 & 55.23 & 56.24 & 64.54 & 1.66 & 78.71 \\
    \specialrule{\lightrulewidth}{0.25em}{0.25em}
    \textbf{Ovis-Embedding-Omni-3B} & \textbf{1} & \textbf{57.29} & \textbf{61.22} &
      \textbf{67.62} & 78.67 & \textbf{61.37} & 56.26 & 66.31 & \textbf{17.59} & 80.70 \\
    \addlinespace[0.15em]
    \bottomrule
  \end{tabular}%
  }
\end{table*}

\paragraph{MAEB results.}
As shown in Table~\ref{tab:maeb-hf-top10}, Ovis-Embedding-Omni-3B achieves the
highest Mean(Task) score of \textbf{57.29} and Mean(Type) score of
\textbf{61.22}. These results exceed the strongest competing scores by 3.75
and 4.10 points, respectively, and place the model first under the estimated
Borda ranking. The gain is broad rather than confined to one task family:
Ovis-Embedding-Omni-3B obtains the best pair-classification (67.62), retrieval (61.37), and
clustering (17.59) results among the models shown. In particular, it improves
retrieval by 5.30 points and clustering by 4.94 points over the corresponding
runners-up. It is also competitive on multilabel and zero-shot
classification, while audio classification and reranking remain below the
best specialized results. Overall, the strong task- and type-level averages
indicate balanced audio representations across both semantic matching and
acoustic discrimination settings.


\begin{table*}[t]
  \centering
  \caption{\textbf{Comparison with the top-ranked models on MVEB(beta).}
  Scores are percentages and are aggregated according to the official task types.
  Rows are ordered by Mean(Task) from low to high. ``Rank'' denotes the
  Borda rank after inserting the locally evaluated
  \textbf{Ovis-Embedding-Omni-3B} results. Best results among the models shown are in
  \textbf{bold}. V-ZS = Video Zero-shot Classification; V-Clf = Video
  Classification; Retr = Any-to-Any Retrieval; V-Clust = Video Clustering;
  V-PC = Video Pair Classification; V-QA = Video-Centric Question Answering.}
  \label{tab:mveb-hf-top10}
  \small
  \setlength{\tabcolsep}{4pt}
  \renewcommand{\arraystretch}{1.22}
  \resizebox{\textwidth}{!}{%
  \begin{tabular}{lccccccccc}
    \toprule
    \multicolumn{1}{c}{\multirow{2}{*}{\textbf{Model}}} &
      & \multicolumn{2}{c}{\textbf{Overall}} &
      \multicolumn{6}{c}{\textbf{TaskType Aggregation}} \\
    \cmidrule(lr){3-4}\cmidrule(l){5-10}
    &
    \textbf{Rank} &
    \textbf{Mean(Task)} & \textbf{Mean(Type)} &
    \textbf{V-ZS} & \textbf{V-Clf} & \textbf{Retr} & \textbf{V-Clust} &
    \textbf{V-PC} & \textbf{V-QA} \\
    \midrule
    BidirLM-Omni-2.5B-Embedding & 9 & 51.20 & 52.05 &
      53.95 & 61.24 & 45.48 & 17.11 & 78.75 & 55.80 \\
    pe-av-small & 11 & 52.18 & 45.64 &
      35.43 & 54.78 & 57.64 & 22.91 & 75.31 & 27.80 \\
    OmniEmbed-v0.1 & 7 & 52.92 & 50.89 &
      48.49 & 58.45 & 52.88 & 20.63 & 74.67 & 50.20 \\
    pe-av-base & 10 & 53.09 & 46.31 &
      34.56 & 54.84 & 59.58 & 22.58 & 75.74 & 30.60 \\
    pe-av-large & 8 & 54.25 & 46.81 &
      37.89 & 54.82 & 62.00 & 21.71 & 75.25 & 29.20 \\
    LCO-Embedding-Omni-3B & 4 & 54.56 & 54.51 &
      52.74 & 56.13 & 54.08 & 27.01 & 80.70 & 56.40 \\
    e5-omni-7B & 3 & 54.99 & 51.17 &
      50.18 & 54.73 & 59.44 & 21.54 & 77.32 & 43.80 \\
    ebind-audio-vision & 5 & 55.46 & 50.27 &
      61.06 & 51.34 & 62.30 & 20.10 & 75.39 & 31.40 \\
    ebind-full & 5 & 55.46 & 50.27 &
      61.06 & 51.34 & 62.30 & 20.10 & 75.39 & 31.40 \\
    LCO-Embedding-Omni-7B & 2 & 57.58 & 56.23 &
      55.46 & 59.23 & 58.71 & \textbf{27.35} & 79.63 & 57.00 \\
    \specialrule{\lightrulewidth}{0.25em}{0.25em}
    \textbf{Ovis-Embedding-Omni-3B} & \textbf{1} & \textbf{61.77} & \textbf{59.72} &
      \textbf{63.94} & \textbf{63.61} & \textbf{63.53} & 25.35 & \textbf{83.30} & \textbf{58.60} \\
    \addlinespace[0.15em]
    \bottomrule
  \end{tabular}%
  }
\end{table*}

\paragraph{MVEB results.}
Table~\ref{tab:mveb-hf-top10} shows a similarly consistent advantage on video understanding. Ovis-Embedding-Omni-3B reaches
\textbf{61.77} Mean(Task) and \textbf{59.72} Mean(Type), outperforming the
second-best model, LCO-Embedding-Omni-7B, by 4.19 and 3.49 points despite its
smaller parameter count. It achieves the best score in five of the six task
types: zero-shot classification (63.94), video classification (63.61),
retrieval (63.53), pair classification (83.30), and video QA (58.60). Video
clustering is the only exception, where its score of 25.35 is 2.00 points
below the best result. This profile suggests that the learned representation
captures both global video semantics and cross-modal audio-visual
correspondence, while leaving some room for improvement in unsupervised
category structure.

\subsubsection{RTEB \& MMEB-Text}
\label{sec:exp-text}
Finally, we compare with more text-specific model to verify where a \emph{omni} model can behave well on broader text search scenarios. Table~\ref{tab:break-text} evaluates text retrieval on two complementary
benchmarks. RTEB targets retrieval in specialised, real-world domains;
we report its 15-task English public split and group the task scores into
legal, code, healthcare, and finance domains. MMEB-Text provides a broader
evaluation over 53 text-retrieval tasks drawn from seven benchmark
families, for which we report only the overall score. Ovis-Embedding-Omni-3B achieves the
best overall results on both RTEB ($67.35$) and MMEB-Text ($47.15$),
outperforming all compared omni-modal and vision-language embedding
models. Notably, it also slightly surpasses Qwen3-Embedding-4B, a larger
model dedicated to text embedding, which scores $67.27$ on RTEB and
$46.22$ on MMEB-Text. These results show that the broader modality coverage
of Ovis-Embedding-Omni-3B does not compromise its text retrieval quality.

\begin{table*}[t]
\centering
\small
\caption{\textbf{RTEB and MMEB-Text results.} RTEB results are reported
  on the 15-task English public split and grouped by domain; its average
  is computed over all 15 tasks. MMEB-Text reports the overall mean over
  53 tasks. The best result in each column is in \textbf{bold}, and the
  second-best is \underline{underlined}.}
\label{tab:break-text}
\setlength{\tabcolsep}{5pt}
\begin{tabular}{l@{\hspace{16pt}}c@{\hspace{20pt}}cccccc}
\toprule
\multicolumn{1}{c@{\hspace{16pt}}}{\multirow{2}{*}{\textbf{Model}}} &
\multirow{2}{*}{\textbf{Size}} &
\textbf{MMEB-Text} & \multicolumn{5}{c}{\textbf{RTEB}} \\
\cmidrule(r){3-3} \cmidrule{4-8}
& & \textbf{Overall} & \textbf{Overall} & \textbf{Legal} &
\textbf{Code} & \textbf{Healthcare} & \textbf{Finance} \\
\midrule
jina-embeddings-v5-omni-small & 1.6B & 45.38 & 63.42 & 54.09 & 66.05 & 64.06 & 66.15 \\
Qwen3-VL-Embedding-2B      & 2.1B & 39.20 & 61.48 & \underline{54.98} & 62.85 & 62.16 & 64.30 \\
LCO-Embedding-Omni-3B      & 3B   & 35.23 & 51.82 & 40.35 & 56.11 & 51.00 & 53.79 \\
Qwen3-Embedding-4B         & 4B   & \underline{46.22} & \underline{67.27} & \textbf{62.67} & \underline{68.47} & \textbf{66.22} & \underline{69.75} \\
\textbf{Ovis-Embedding-Omni-3B} & 3B & \textbf{47.15} & \textbf{67.35} & 49.61 & \textbf{69.89} & \underline{66.13} & \textbf{80.00} \\
\bottomrule
\end{tabular}
\end{table*}

\subsection{VL-9B \& VL-2B Results}
\label{sec:vl-9b-2b-results}

\begin{table*}[t]
  \centering
  \caption{\textbf{Results on MMEB-v2.}
  We compare Ovis-Embedding-VL-9B with representative vision--language
  embedding models across image, video, and visual-document retrieval tasks.
  The baselines include seed1.6-embedding-1215~\citep{seed1_6_embedding_1215},
  Qwen3-VL-Embedding-8B~\citep{qwen3_vl_embedding},
  DME-Medium~\citep{dme}, and
  Octen-VL-Embedding-Large~\citep{octen_vl}.
  Scores are reported as percentages. Overall is the unweighted average over
  all 78 constituent datasets. The best and second-best results in each row
  are shown in \textbf{bold} and \underline{underlined}, respectively.}
  \label{tab:mmeb-v2}
  \small
  \setlength{\tabcolsep}{6pt}
  \renewcommand{\arraystretch}{1.04}
  \renewcommand{\multirowsetup}{\centering}
  \begin{tabular*}{\textwidth}{@{}>{\centering\arraybackslash}m{0.11\textwidth}@{\hspace{4pt}}>{\centering\arraybackslash}m{0.16\textwidth}@{\hspace{4pt}\extracolsep{\fill}}ccccc@{}}
    \toprule
    \multirow[c]{2}{=}{\raisebox{-1.4\baselineskip}{\textbf{Modality}}} &
    \multirow[c]{2}{=}{\raisebox{-1.4\baselineskip}{\textbf{Sub-task}}} &
    \multicolumn{5}{c}{\textbf{Model}} \\
    \cmidrule(r){3-7}
    & &
    {\footnotesize\shortstack[c]{\textbf{Ovis-VL-}\\\textbf{Embedding-}\\\textbf{9B}}} &
    {\footnotesize\shortstack[c]{seed1.6-\\embedding-\\1215}} &
    {\footnotesize\shortstack[c]{Qwen3-VL-\\Embedding-\\8B}} &
    {\footnotesize\raisebox{0.5\baselineskip}{\shortstack[c]{DME-\\Medium}}} &
    {\footnotesize\shortstack[c]{Octen-VL-\\Embedding-\\Large}} \\
    \midrule

    \multirow[c]{5}{=}{\textbf{IMG}}
      & \textit{Overall} & \textbf{83.96} & 77.99 & 80.12 & 79.76 & \underline{81.86} \\
      & I-CLS            & \textbf{80.11} & 75.05 & 74.19 & 74.48 & \underline{75.68} \\
      & I-QA             & \textbf{86.72} & 74.89 & 81.14 & 80.94 & \underline{84.53} \\
      & I-RET            & \textbf{81.03} & 79.33 & 80.16 & 78.23 & \underline{80.62} \\
      & I-VG             & \textbf{95.47} & 89.05 & 92.30 & \underline{94.55} & 94.40 \\
    \midrule

    \multirow[c]{5}{=}{\textbf{VID}}
      & \textit{Overall} & \underline{72.90} & 67.74 & 67.15 & 70.79 & \textbf{75.95} \\
      & V-CLS            & \textbf{88.41} & 85.19 & 78.39 & \underline{87.69} & 87.09 \\
      & V-QA             & \underline{72.50} & 66.71 & 70.96 & 71.03 & \textbf{82.14} \\
      & V-RET            & \underline{62.61} & 59.09 & 58.73 & 60.99 & \textbf{67.99} \\
      & V-MRET           & \textbf{64.86} & 54.79 & 56.09 & 58.53 & \underline{60.36} \\
    \midrule

    \multirow[c]{5}{=}{\textbf{VISDOC}}
      & \textit{Overall} & \textbf{83.06} & \underline{82.38} & 82.36 & 82.02 & 80.54 \\
      & ViDoRe-V1        & \textbf{91.22} & \underline{90.90} & 87.21 & 87.62 & 85.90 \\
      & ViDoRe-V2        & 61.00 & 60.31 & \textbf{69.86} & 57.79 & \underline{66.34} \\
      & VisRAG           & \underline{91.19} & 89.96 & 88.68 & \textbf{94.50} & 86.64 \\
      & VisDoc-OOD       & 72.50 & 71.76 & \underline{73.27} & \textbf{73.54} & 72.16 \\
    \midrule

    \textbf{ALL} & \textit{Overall} & \textbf{81.13} & 76.97 & 77.82 & 78.38 & \underline{80.09} \\

    \bottomrule
  \end{tabular*}
\end{table*}

Having established the omni-modal performance of Ovis-Embedding-Omni-3B, we now turn to the vision--language variants and evaluate the proposed approach at two model scales. Specifically, we report the MMEB-v2 results of Ovis-Embedding-VL-2B and Ovis-Embedding-VL-9B, focusing on their overall performance, modality-specific strengths, and scaling behavior.

Tables~\ref{tab:mmeb-v2} and~\ref{tab:mmeb-v2-2b} compare the 9B and 2B
variants of Ovis-Embedding-VL with representative models of similar scale on
MMEB-v2. Both variants achieve the best aggregate score within their
respective comparison groups. Ovis-Embedding-VL-9B reaches an overall score
of 81.13, outperforming the strongest selected baseline,
Octen-VL-Embedding-Large, by 1.04 points. Ovis-Embedding-VL-2B obtains 77.46,
exceeding Octen-VL-Embedding by 2.04 points. Since the overall metric averages
all 78 constituent datasets rather than the three modality-level scores, these
gains indicate broadly balanced improvements rather than dominance on only a
small subset of tasks.

\paragraph{Image tasks.}
The clearest advantage of both variants appears on image tasks. The 9B model
achieves an IMG average of 83.96, improving upon the next-best result by 2.10
points, while the 2B model obtains 80.62 with a larger margin of 3.21 points.
Both variants rank first on all four image sub-tasks: classification,
question answering, retrieval, and visual grounding. Overall, these
results demonstrate particularly strong visual semantic discrimination and
question-guided matching, together with consistently competitive performance
across all four image sub-tasks.

\paragraph{Video tasks.}
On the more competitive video suite, Ovis-Embedding-VL-9B and
Ovis-Embedding-VL-2B achieve VID averages of 72.90 and 67.12, respectively.
Both variants perform particularly strongly on tasks that require temporal
localization: each ranks first on video moment retrieval, with scores of
64.86 and 57.63 for the 9B and 2B models. Both variants also obtain the best
video classification result in their respective comparison groups.
Performance on the remaining video tasks is competitive and suggests further
potential for strengthening broad temporal-semantic matching.

\paragraph{Visual-document tasks.}
On visual-document tasks, Ovis-Embedding-VL-9B and
Ovis-Embedding-VL-2B achieve the best modality-level averages in their
respective comparison groups, with scores of 83.06 and 80.47. Both variants
also lead on ViDoRe-V1, while the 2B model obtains the strongest result on
out-of-distribution document retrieval. Taken together, the results show
robust document representations across both model sizes, while multilingual
and domain-diverse document retrieval offers a promising direction for
further improvement.

\paragraph{Scaling behavior.}
Comparing the two Ovis variants directly further highlights the effect of
model scale. Moving from 2B to 9B improves the overall score by 3.67 points,
with gains of 3.34 on IMG, 5.78 on VID, and 2.59 on VISDOC. The largest
sub-task improvements occur on video question answering ($+7.64$), video
moment retrieval ($+7.23$), and video retrieval ($+5.74$). Thus, additional
capacity benefits all three modalities, but contributes most strongly to
temporal understanding and matching, which require integrating information
across longer and more complex visual sequences.
\begin{table*}[t]
  \centering
  \caption{\textbf{Results on MMEB-v2.}
  We compare Ovis-Embedding-VL-2B with representative vision--language
  embedding models across image, video, and visual-document retrieval tasks.
  The baselines include UEmbed-2B (dense)~\citep{uembed2026},
  Qwen3-VL-Embedding-2B~\citep{qwen3_vl_embedding},
  DME-Small~\citep{dme}, and Octen-VL-Embedding~\citep{octen_vl}.
  Scores are reported as percentages. Overall is the unweighted average over
  all 78 constituent datasets. The best and second-best results in each row
  are shown in \textbf{bold} and \underline{underlined}, respectively.}
  \label{tab:mmeb-v2-2b}
  \small
  \setlength{\tabcolsep}{6pt}
  \renewcommand{\arraystretch}{1.04}
  \renewcommand{\multirowsetup}{\centering}
  \begin{tabular*}{\textwidth}{@{}>{\centering\arraybackslash}m{0.11\textwidth}@{\hspace{4pt}}>{\centering\arraybackslash}m{0.16\textwidth}@{\hspace{4pt}\extracolsep{\fill}}ccccc@{}}
    \toprule
    \multirow[c]{2}{=}{\raisebox{-1.4\baselineskip}{\textbf{Modality}}} &
    \multirow[c]{2}{=}{\raisebox{-1.4\baselineskip}{\textbf{Sub-task}}} &
    \multicolumn{5}{c}{\textbf{Model}} \\
    \cmidrule(r){3-7}
    & &
    {\footnotesize\shortstack[c]{\textbf{Ovis-Embedding-}\\\textbf{VL-2B}}} &
    {\footnotesize\shortstack[c]{UEmbed-2B\\(dense)}} &
    {\footnotesize\shortstack[c]{Qwen3-VL-\\Embedding-2B}} &
    {\footnotesize\shortstack[c]{DME-\\Small}} &
    {\footnotesize\shortstack[c]{Octen-VL-\\Embedding}} \\
    \midrule

    \multirow[c]{5}{=}{\textbf{IMG}}
      & \textit{Overall} & \textbf{80.62} & 67.76 & 74.96 & 75.91 & \underline{77.41} \\
      & I-CLS            & \textbf{76.87} & 59.21 & 70.34 & 68.63 & \underline{71.49} \\
      & I-QA             & \textbf{82.05} & 67.91 & 74.27 & 76.21 & \underline{78.90} \\
      & I-RET            & \textbf{77.69} & 68.81 & 74.85 & 75.59 & \underline{76.37} \\
      & I-VG             & \textbf{95.17} & 85.58 & 88.55 & \underline{94.30} & 91.58 \\
    \midrule

    \multirow[c]{5}{=}{\textbf{VID}}
      & \textit{Overall} & \underline{67.12} & 49.96 & 61.87 & 65.63 & \textbf{68.84} \\
      & V-CLS            & \textbf{85.32} & 56.26 & 71.94 & \underline{84.46} & 79.08 \\
      & V-QA             & 64.86 & 52.76 & \underline{64.94} & 61.90 & \textbf{73.61} \\
      & V-RET            & \underline{56.87} & 45.76 & 53.88 & 55.48 & \textbf{61.34} \\
      & V-MRET           & \textbf{57.63} & 41.78 & 53.26 & \underline{57.38} & 56.33 \\
    \midrule

    \multirow[c]{5}{=}{\textbf{VISDOC}}
      & \textit{Overall} & \textbf{80.47} & 76.99 & 79.22 & \underline{79.86} & 77.38 \\
      & ViDoRe-V1        & \textbf{88.87} & 82.31 & 84.37 & \underline{86.99} & 82.45 \\
      & ViDoRe-V2        & 57.08 & 59.89 & \textbf{65.34} & 56.24 & \underline{63.03} \\
      & VisRAG           & \underline{89.05} & 85.60 & 86.43 & \textbf{90.32} & 84.14 \\
      & VisDoc-OOD       & \textbf{70.02} & 67.90 & 69.41 & \underline{69.98} & 68.87 \\
    \midrule

    \textbf{ALL} & \textit{Overall} & \textbf{77.46} & 66.49 & 73.25 & 74.75 & \underline{75.42} \\

    \bottomrule
  \end{tabular*}
\end{table*}

\section{Related Work}
\label{sec:related}
We situate Ovis-Embedding in three complementary lines of prior research. The first concerns dense text embedding models, which establish the contrastive recipe and the prompt-based pooling strategies that we inherit. The second covers multimodal and unified embedding models, which extend such recipes beyond text but typically remain centered on the image--text pair. The third surveys multimodal large language models, whose hidden states have recently emerged as a strong backbone for embedding extraction and which provide the immediate point of departure for Ovis-Embedding.

\paragraph{Text Embedding Models.}
Dense text embeddings have progressed rapidly from static word
vectors through sentence transformers to LLM-based encoders.
DPR~\citep{dpr} popularized dual-encoder contrastive training for
open-domain question answering.  E5~\citep{e5} and GTE~\citep{gte}
demonstrated that a single model can span classification, clustering,
retrieval and semantic similarity tasks when trained with a large-scale
weakly-supervised contrastive corpus and a multi-stage recipe.
BGE-M3~\citep{bge_m3} further pushed the frontier by simultaneously
supporting multi-linguality, multi-functionality (dense, sparse, and
multi-vector retrieval), and multi-granularity (up to 8k tokens),
achieving state-of-the-art results across multiple language families.
More recently, LLM-based embedding models such as
E5-Mistral~\citep{e5mistral} and NV-Embed~\citep{nvembed} have shown
that decoder-only architectures can serve as generalist embedding
backbones when equipped with instruction-based task prompts and
last-token pooling, achieving state-of-the-art results on the
MTEB benchmark~\citep{mteb}.  Our training recipe inherits many ideas
from this line---prompt-based pooling, multi-stage contrastive
pretraining, on-policy hard-negative mining~\citep{ance}---and
extends them from a text-only setting to a native omni-modal one.

\paragraph{Multimodal Embedding Models.}
Extending dense retrieval beyond text has primarily been explored
through CLIP~\citep{clip} and its successors
(SigLIP~\citealp{siglip}, Jina-CLIP~\citealp{jina_clip}), which train
dual-encoder image--text models on web-scale data.
A series of follow-up works~\citep{pace} have repurposed vision--language models
(VLMs) for embedding extraction.
VLM2Vec~\citep{vlm2vec} finetunes VLMs on the MMEB benchmark and
achieves 10--20\% absolute gains over prior approaches; its recent
successor VLM2Vec-V2~\citep{vlm2vec_v2} scales this paradigm to
larger backbones with improved training data.
GME~\citep{gme} builds on Qwen2-VL~\citep{qwen2vl} and proposes
unified multimodal retrieval that supports text-only, image-only
and interleaved text--image queries.
Qwen3-VL-Embedding~\citep{qwen3_vl_embedding} extends the Qwen3-VL
family to embedding tasks, achieving state-of-the-art performance on
vision-language benchmarks (MMEB, ViDoRe) but without native audio
support.

The most recent wave of work has pursued \emph{omni-modal}
embeddings that cover text, image, video, and audio within a
single model.
LCO-Embedding-Omni~\citep{lco_embedding} introduces learned
compression tokens to unify modalities into a compact embedding
space, reporting strong audio results but weaker vision scores.
e5-omni~\citep{e5_omni} adapts a multimodal LLM for omni-modal
embeddings via modality-aware contrastive learning.
Conan-embedding-v3~\citep{conan_embedding} leverages large-scale
negative mining and curriculum training to achieve leading
performance on both MMEB and MAEB benchmarks.
jina-embeddings-v5~\citep{jina_v5_omni} proposes a multi-task
architecture supporting text, image, and audio embeddings.
In the audio domain, CLAP~\citep{clap} applies contrastive
language--audio pretraining to align audio and text in a shared space,
but remains limited to the audio--text pair.

A common limitation shared by these recent omni-modal systems is
that they either (i) rely on separate modality adapters bolted onto
a text backbone, or (ii) sacrifice performance on one modality
(typically vision) to accommodate audio. Our work departs from this
pattern by adopting Qwen-Omni as backbone---a model that
natively processes text, image, audio, and video within a single
Thinker module---enabling genuine joint fusion rather than late-stage
modality concatenation.

\paragraph{Benchmarks.}
The Massive Text Embedding Benchmark (MTEB)~\citep{mteb} established
a standardized evaluation for text embeddings spanning eight task
categories.
VLM2Vec~\citep{vlm2vec} introduced Massive Multimodal
Embedding Benchmark (MMEB), a 36-dataset suite that extends MTEB-style
evaluation to multimodal inputs covering classification, VQA,
retrieval and visual grounding.
In this paper we evaluate our models on these benchmarks, complemented by the
Massive Audio Embedding Benchmark (MAEB)~\citep{maeb} and the
Massive Video Embedding Benchmark (MVEB)~\citep{mveb} for focused
evaluation of audio and video representations. Full details of all benchmark configurations are given in Appendix~\ref{app:results}.

\section{Conclusion}
\label{sec:conclusion}
We presented Ovis-Embedding, a universal embedding family combining native multimodal backbones, data-centric training, and embedding-specific optimization. The family achieves state-of-the-art performance on MMEB-v3, MMEB-v2, MAEB, and MVEB, with flexible embedding dimensions for efficient deployment. These results motivate further exploration of native omni-modal models as a foundation for universal retrieval.


\section*{Authors and Contributions}
\label{sec:authors}
\paragraph{Core Contributors.}
Ke Zhu, Yawen Liu, Wenjun Yan, Yuwei Hu, Xin Wang

\paragraph{Contributors.}
Jin Tong, Wei Zhou, Yibo Wang, Liwei Liu, Dawei Yan, Yanping Li, Jiaming Chen

\paragraph{Project Leaders.}
Guangda Huzhang, Qing-Guo Chen, Zhao Xu, Weihua Luo

\bibliography{references}
\bibliographystyle{colm2024_conference}

\appendix
\clearpage

\label{app:impl}



\section{Additional Experimental Results}
\label{app:results}

\paragraph{Elastic Embedding Results.}
Table~\ref{tab:elastic-dims} (in Appendix) reports all six suites at the five nested
widths, with the averages weighted by the number of tasks in each suite according to MMEB-v3 tasks. Halving the
embedding to $d=1024$ is free ($100.1\%$ retention), $d=512$ and $d=256$ cost
$0.8\%$ and $2.6\%$ of it, and $d=128$ --- a sixteen-fold reduction --- still
retains $93.2\%$. The baseline throughout
is naive truncation: the same frozen embedding cut to its first $d$
coordinates and renormalized. Our
margin over it grows monotonically as the prefix shortens,
from $+0.43$ at $d=1024$ to $+4.30$ at $d=128$. At $d=1024$ three suites sit
marginally above full width, by at most $0.32$; we read these as no
measurable change rather than as gains, and as an indication that little is
left to recover at that width --- the useful range of the adaptation is the
short prefixes.

The average also hides a wide spread across suites
(c.f.~Fig.~\ref{fig:curves}). At $d=128$
audio loses $0.33$ points and video $1.26$, whereas VisDoc and Agent lose
$6.97$ and $6.18$. Plain truncation costs those two the most as well,
$13.89$ and $11.01$ points against $3.00$ on audio, and they are also where
the adaptation recovers the least of that loss: the gap is a
property of the task rather than of the compressor. We attribute this to the limited information capacity of a low-dimensional vector. Audio and video
queries here are largely settled by coarse category-level
evidence, which survives in the leading directions, whereas VisDoc has to
localize one passage or figure within a page whose global appearance it
shares with many distractors, and Agent has to separate interface states
that differ in a few small elements. Such fine-grained decisions
rest on lower-variance directions, and those are precisely the directions a
short prefix discards, in which case what these suites need is
capacity, not a better projection.
\begin{table}[t]
\centering
\small
\caption{\textbf{Task performance at five nested widths.} Our full
  pipeline evaluated on the six MMEB-v3 suites. The $2,048$ column is the
  frozen encoder itself rather than a separate run: adapters are fitted
  only for $d < D$, and the PCA basis is orthogonal, so the full-width
  representation is unchanged.
  \emph{Retention} is that average relative to $2,048$; the last two rows
  give naive truncation of the same encoder and our gain over it.}
\label{tab:elastic-dims}
\begin{tabular}{llccccc}
\toprule
\textbf{Task Suite} & \textbf{Metric} &
\textbf{2,048} & \textbf{1,024} & \textbf{512} & \textbf{256} & \textbf{128} \\
\midrule
MMEB-Text        & nDCG@5 & 46.78 & 46.69 & 46.15 & 45.30 & 43.05 \\
MMEB-Image       & Hit@1  & 77.23 & 77.30 & 77.27 & 76.70 & 75.50 \\
MMEB-Video       & Hit@1  & 64.43 & 64.26 & 64.25 & 63.92 & 63.17 \\
MMEB-Audio       & Hit@1  & 48.02 & 48.04 & 48.16 & 48.04 & 47.69 \\
MMEB-VisDoc      & nDCG@5 & 77.73 & 77.60 & 76.85 & 74.64 & 70.76 \\
MMEB-Agent       & Hit@1  & 45.32 & 45.64 & 44.66 & 43.05 & 39.14 \\
\midrule
\textbf{Avg.}  &        & \textbf{58.00} & \textbf{58.04} & \textbf{57.55}
                          & \textbf{56.49} & \textbf{54.08} \\
Retention (\%)   &        & 100.0 & 100.1 & 99.2 & 97.4 & 93.2 \\
\midrule
Naive truncation (Avg.) & & 58.00 & 57.61 & 56.28 & 54.42 & 49.78 \\
Retention (\%) &  & 100.0 & 99.3 & 97.0 & 93.8 & 85.8 \\
\bottomrule
\end{tabular}
\end{table}

\begin{figure}[h]
\centering
\includegraphics[width=\linewidth]{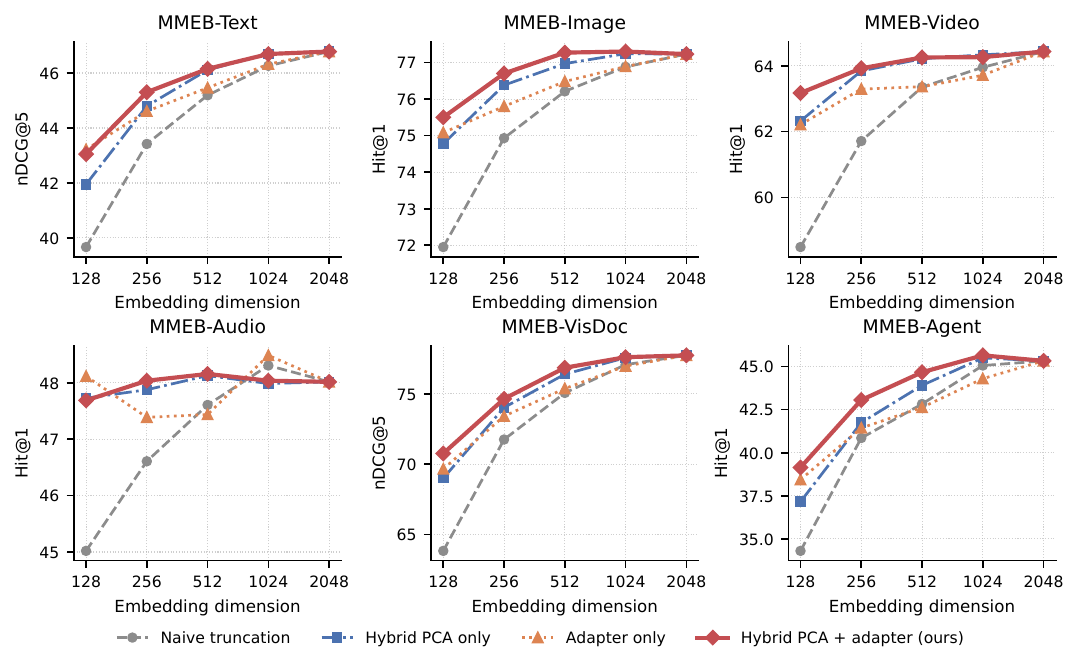}
\caption{\textbf{Performance versus embedding width on all six MMEB-v3
  suites.} We compare naive truncation of
  the frozen encoder, the hybrid PCA basis alone, the residual adapter alone, and the
  two combined. Note
  that the vertical scale and the metric differ per panel (Hit@1 for image,
  video, audio and agent; nDCG@5 for text and VisDoc), so panels show
  trends rather than comparable magnitudes. 
  Two patterns run across the panels. The PCA basis carries most of the gain
  at $512$ and above while the adapter carries more of it at $128$
  (averaged over the six suites, $59.28$ vs.\ $58.46$ at $512$, reversing
  to $55.49$ vs.\ $56.12$ at $128$), and neither component alone reaches
  the combination at any width. The adapter alone is also the only
  configuration that can fall below plain truncation, which it does at
  $1024$ on video, VisDoc and Agent; composing it with the basis removes
  that regression, consistent with the basis supplying the initialization
  the adapter is trained from.}
\label{fig:curves-all}
\end{figure}




\section{Example of Data}
\label{app:data-}
\begin{figure*}[ht]
\centering
\includegraphics[width=1.0\textwidth]{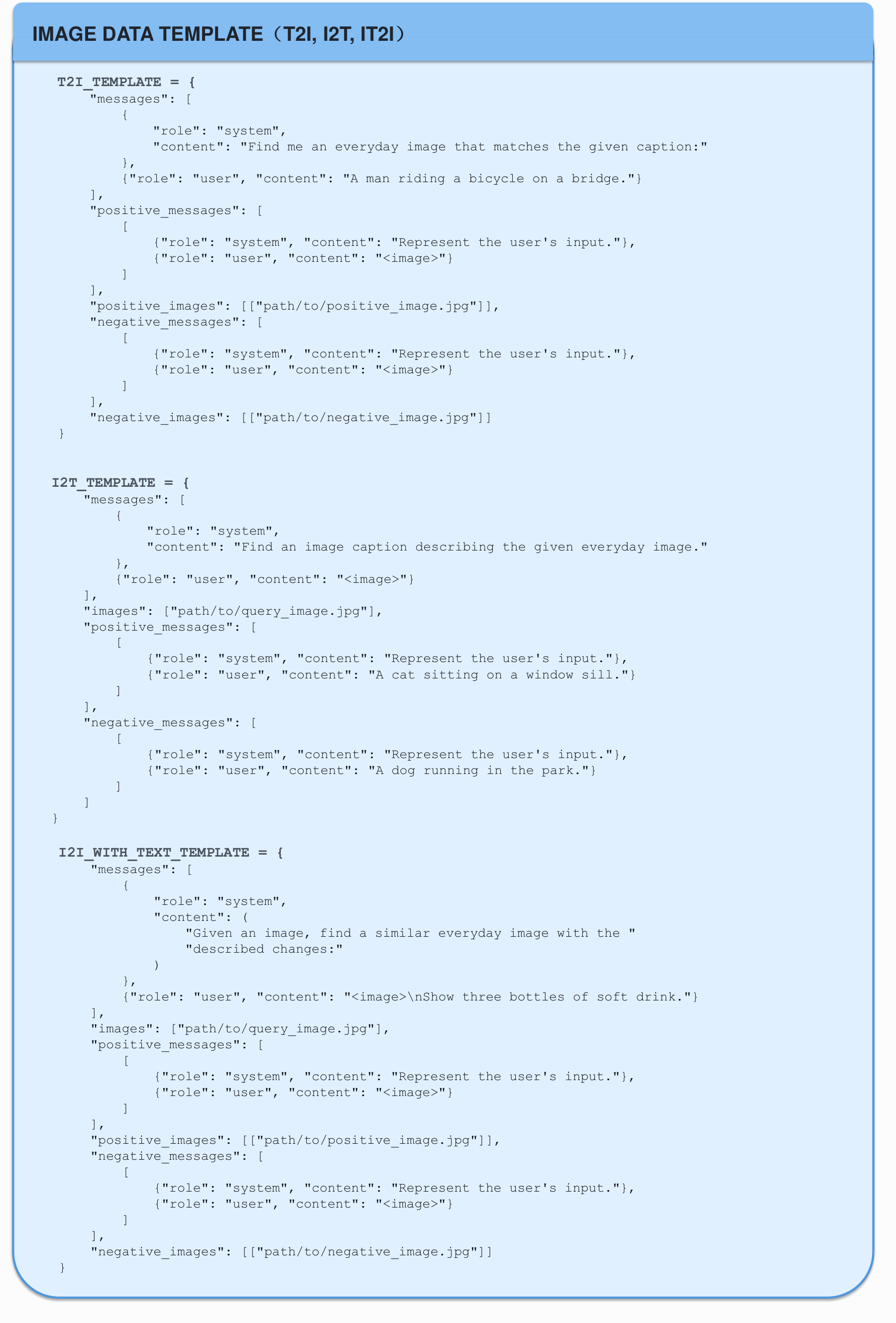}
\caption{Image data template for text-to-image, image-to-text and text-image-to-text.}
\label{fig:image-data-1}
\end{figure*}

\begin{figure*}[ht]
\centering
\includegraphics[width=1.0\textwidth]{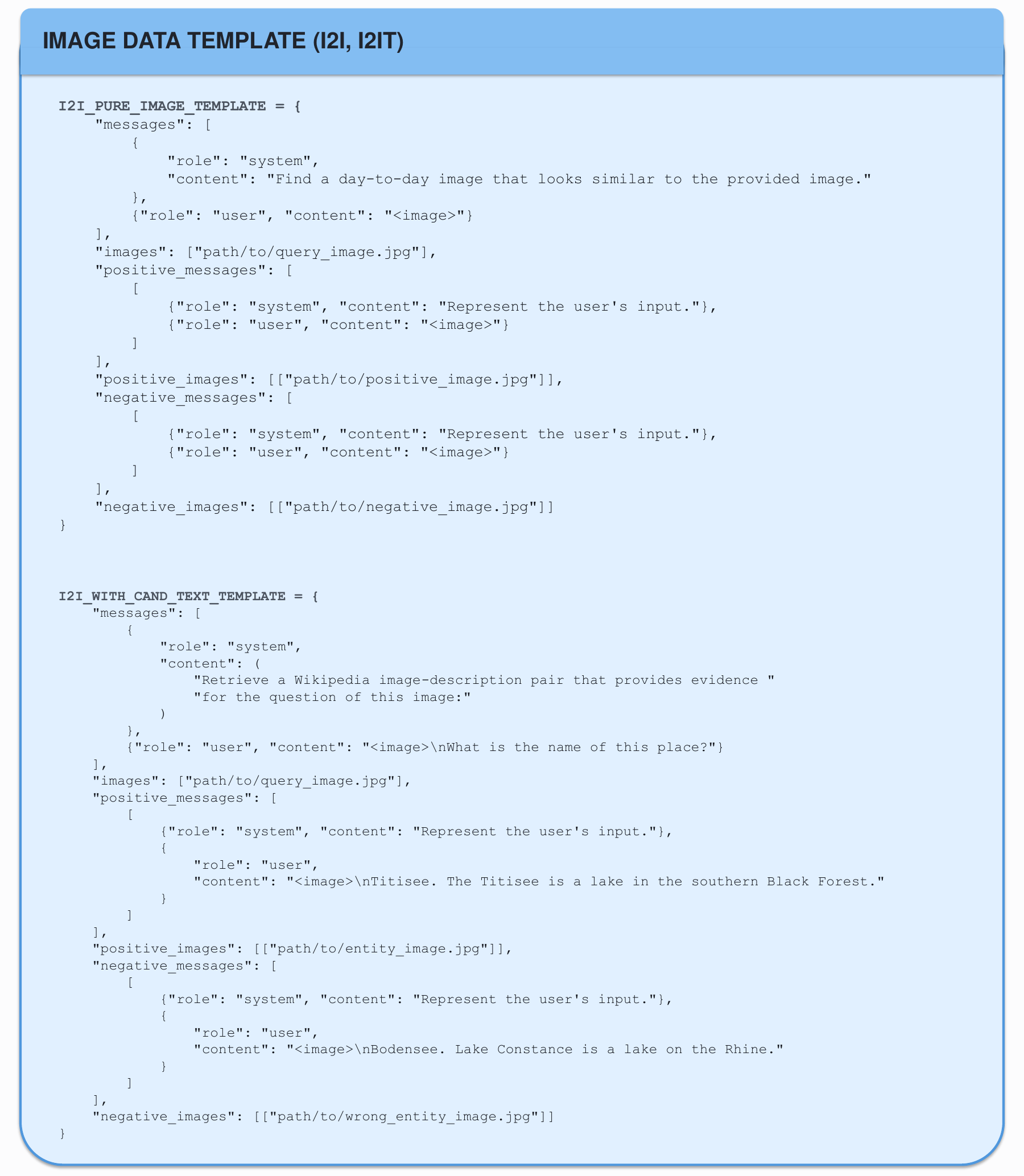}
\caption{Image data template for image-to-image, image-to-image-text.}
\label{fig:image-data-2}
\end{figure*}

\begin{figure*}[t]
\centering
\includegraphics[width=1.0\textwidth]{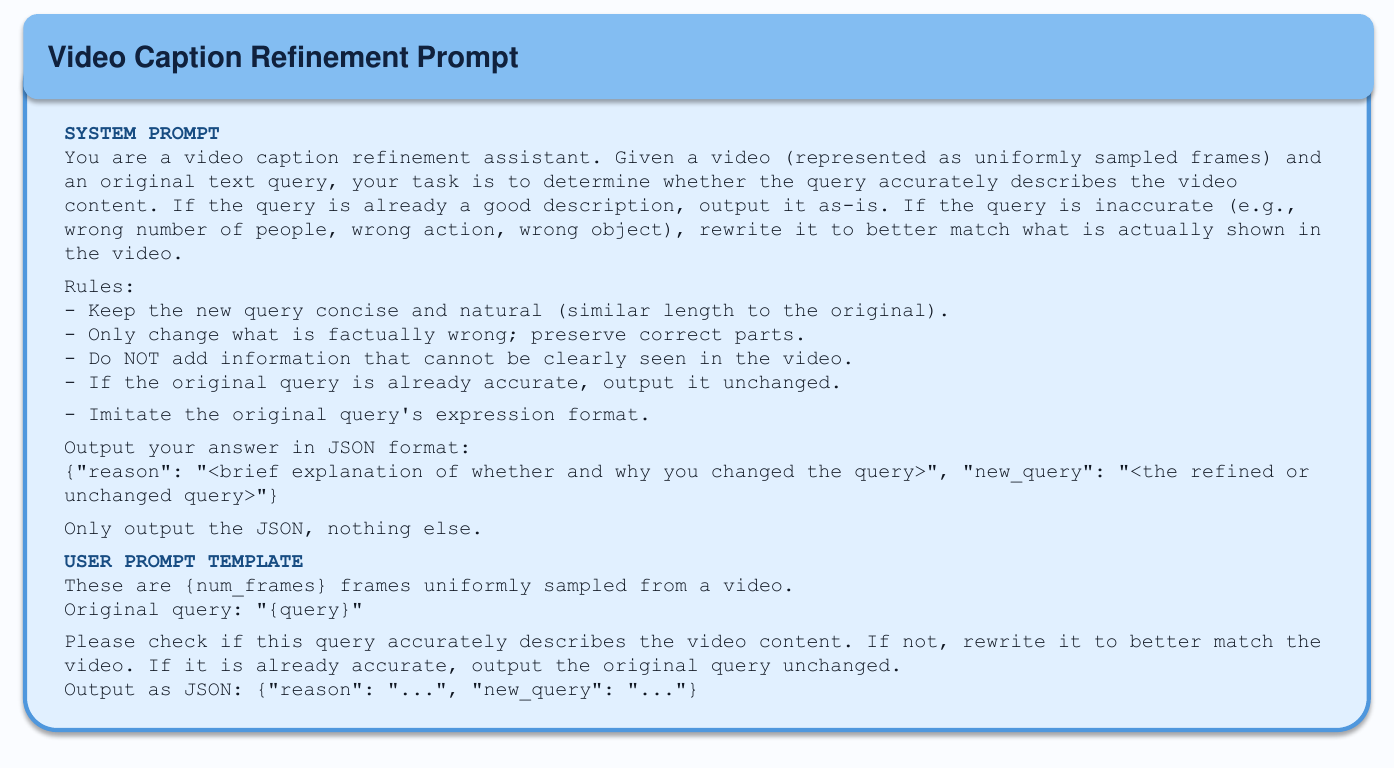}
\caption{Video query rewrite prompt}
\label{fig:video-prompt-1}
\end{figure*}

\begin{figure*}[t]
\centering
\includegraphics[width=1.0\textwidth]{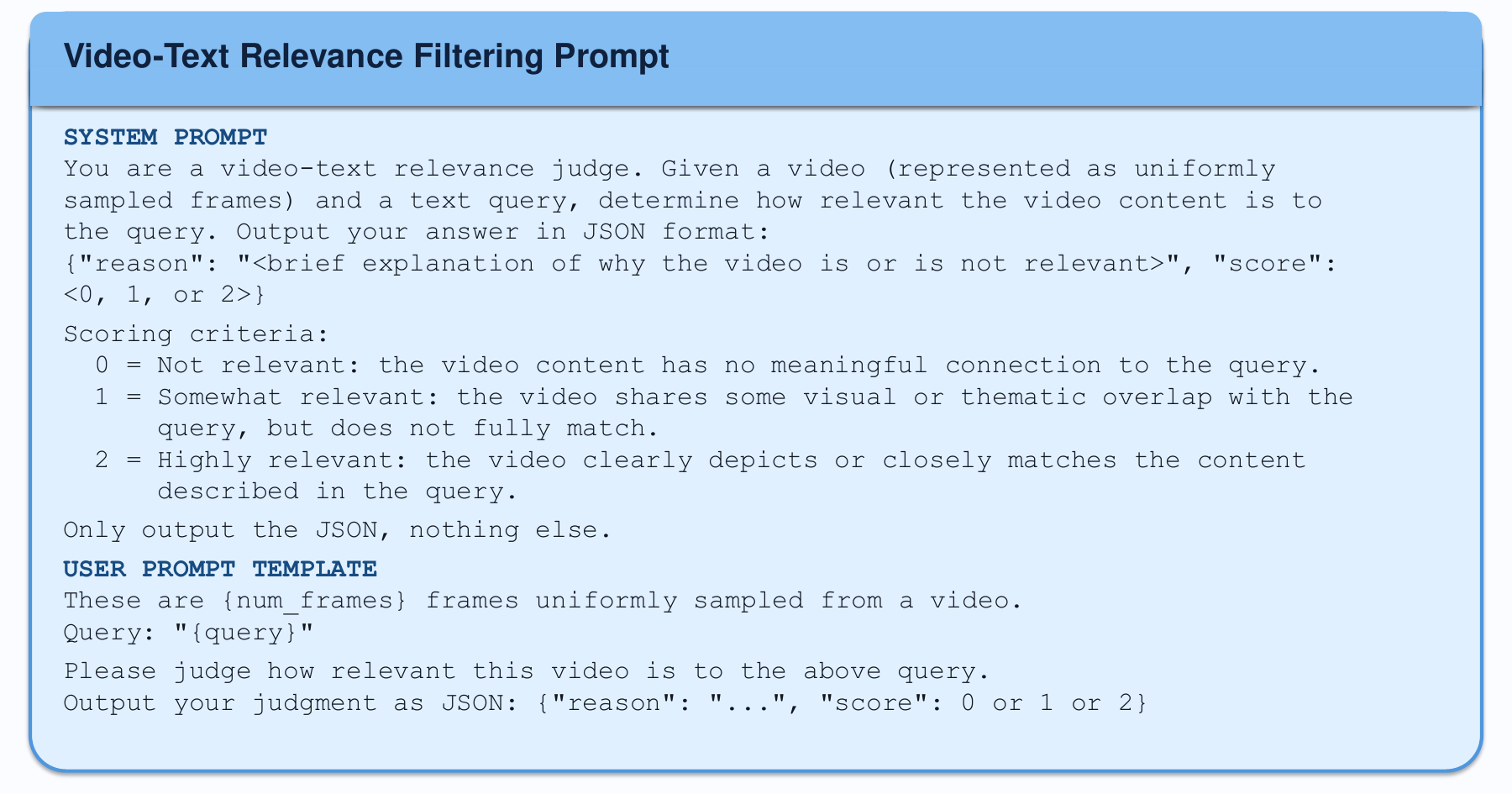}
\caption{Video relevance filtering prompt}
\label{fig:video-prompt-2}
\end{figure*}

 Please refer to Fig.~\ref{fig:image-data-1} and Fig.~\ref{fig:image-data-2} for image data format and refer to Fig.~\ref{fig:video-prompt-1}-\ref{fig:video-prompt-2} for video data synthesis.

\end{document}